\documentclass{article}
\usepackage{helper/packages_and_styles}

\definecolor{pink}{RGB}{200, 0, 100}
\definecolor{mblue}{rgb}{0.2, 0.48, 0.65}

\newcommand{\ie}{\textit{i}.\textit{e}.~}
\newcommand{\eg}{\textit{e}.\textit{g}.~}

\newcommand{\q}{$q$}

\newcommand{\kendall}{Kendall's $\tau_b$}

\usepackage{adjustbox}
\usepackage{array}

\newcolumntype{R}[2]{%
    >{\adjustbox{angle=#1,lap=\width-(#2)}\bgroup}%
    l%
    <{\egroup}%
}
\newcommand{\sectioncolor}{mblue} % for coloring sections in datasheet

\newcommand{\datasetDOI}{\url{https://figshare.com/s/65bfb8f90a454207a275}} % 10.17044/scilifelab.14687271
\newcommand{\datasetURL}{URL}

\newcommand{\maskingRepo}{\url{https://github.com/yueliukth/CSAW-M/}}

\newcommand{\SciRepo}{https://scilifelab.figshare.com/}
\usepackage{helper/packages_and_styles}

\usepackage{listings}
\usepackage{xcolor}

\title{CSAW-M: An Ordinal Classification Dataset for Benchmarking Mammographic Masking of Cancer}
\author[1, 2]{Moein Sorkhei\thanks{Equal contribution}\hspace{1.5mm}}
\author[1, 2]{Yue Liu$^*$}
\author[1]{Hossein Azizpour}
\author[3,4]{\\Edward Azavedo} 
\author[3,5]{Karin Dembrower}
\author[4]{Dimitra Ntoula}
\author[3,4]{\\Athanasios Zouzos}
\author[3,4]{Fredrik Strand}
\author[1, 2]{Kevin Smith}

\affil[1]{KTH Royal Institute of Technology, Stockholm, Sweden}
\affil[2]{SciLifeLab, Stockholm, Sweden}
\affil[3]{Karolinska Institutet, Stockholm, Sweden}
\affil[4]{Karolinska University Hospital, Stockholm, Sweden}
\affil[5]{Saint Göran Hospital, Stockholm, Sweden}

\begin{document}

\maketitle

%\maketitle
\begin{abstract}
%Interval and large invasive breast cancers, often caused by mammographic masking potential when the lesion is obscured, are commonly detected at a later stage and associated with a worse prognosis. To study and benchmark mammographic masking of cancer, we introduce the \textit{CSAW-M}, a large-scale ordinal classification dataset with valuable masking assessments from experts. In contrast to previous approaches measuring breast density as a proxy, our dataset is a direct measure of masking potential. We developed a deep learning model that is capable of estimating the masking level, and have shown that the predicted masking estimate performs better in identifying women diagnosed with interval and large invasive cancers, without being explicitly trained for these tasks, than its breast density counterparts.
Interval and large invasive breast cancers, which are associated with worse prognosis than other cancers, are usually detected at a late stage due to false negative assessments of screening mammograms. 
The missed screening-time detection is commonly caused by the tumor being obscured by its surrounding breast tissues, a phenomenon called masking.  
To study and benchmark mammographic masking of cancer, in this work we introduce CSAW-M, the largest public mammographic dataset, collected from over 10,000 individuals and annotated with potential masking. 
In contrast to the previous approaches which measure breast image density as a proxy, our dataset directly provides annotations of masking potential assessments from five specialists. 
We also trained deep learning models on CSAW-M to estimate the masking level and showed that the estimated masking is significantly more predictive of screening participants diagnosed with interval and large invasive cancers -- without being explicitly trained for these tasks -- than its breast density counterparts.
%\ha{I mainly redid what Yue wrote but it might be good to mention that CSAW-M has CEP annotations and that we plan to provide a private benchmark. Also, I am not sure if it's a good idea to have a footnote in the abstract}
%Colors: \ms{Moein}, \ks{Kevin}, \yl{Yue}, \fs{Fredrik}, \ha{Hossein}. \ks{We should analyze the correlation between density measures and masking, either in this paper or another clinical paper led by KI, ask Fredrik.}
\end{abstract}
% Fredrik's suggestion
% When breast cancers escape detection in mammography screening programs, the affected women suffer a higher risk of dying since the cancer is given more time to grow and spread beyond the local environment. Missed screen-detection is commonly caused by surrounding high-density breast tissue obscuring the tumor, a phenomenon called masking. To study and benchmark masking, in this work we introduce CSAW-M, the largest public mammography dataset, collected from over 10,000 individuals and annotated by five radiologists according to the potential of masking. In contrast to this direct assessment of masking, traditional approaches have been entirely or partly based on using the proportion of high-density pixels as a proxy. We combined binary image decisions by the radiologists with a sorting mechanism to assign an ordinal level of masking for each image. We trained deep learning models on CSAW-M to produce a masking estimator. We showed that the estimated masking is significantly more predictive of women whose cancer was potentially missed at screening than the traditional breast density approach.

\section{Introduction}

%\yl{What we want to include in this section}
%\begin{itemize}
        % \item \yl{Clinical motivation and importance of masking} 
        % \item \yl{Reasons why tails are interesting in clinical studies } \ms{Why not only high-masking is important, but also low-masking is important and interesting for us (e.g. with respect to resources in a hospital and the fact that low-masking one may. not need more advance imaging technology))}
        % \item \yl{Show examples of images in level1-8 - images all or most radiologists agree on in each level; later on we show images that the radiologists couldn't reach consensus on}
 %       \item \yl{We may also need a table to compare our dataset with other datasets such as CBIS-DDSM} \ms{Comparing different properties that relevant mammography have or what they miss (e.g. wehter film/digital, their image size, their number of images, whether they have BIRADS, etc. (and possibly visualized images of how different their images are)¨} \yl{maybe in appendix?} \ks{probably appendix, but mention in text... actually on second thought, this would be good to include at the end of the intro or in the related work.}
        % \item \ms{Definition of (percent) density and masking.} 
        % \item first indicating hard images (hard to interpret, needing more advanced imaging technologies), second indicating low images (which need less attention)
%\end{itemize}

Regular mammographic screening helps detect breast cancer at an early stage, and has been demonstrated to decrease mortality by around 30\% \cite{tabar2011swedish}. 
However, 17-30\% of breast cancers among screening participants are \emph{interval cancers} -- cancers detected clinically after a negative screening \cite{houssami2017epidemiology}.
Interval cancers are often associated with a worse prognosis \cite{meshkat2015comparison, eriksson2013mammographic}.
So-called \emph{true interval cancers} are characterized by rapid growth after a healthy mammogram, while \emph{missed interval cancers} are the result of false-negative assessments of a mammogram, often because the lesion is obscured or \emph{masked} by breast tissue.

%\yl{to do - motivate large invasive cancer}%\ha{In this work, we introduce a dataset of mammograms annotated, by expert radiologists, with different levels of masking.} \ks{[We say almost the same thing in 4th paragraph]}

%Apart from true interval cancers with rapid tumor growth patterns developed after a prior negative mammogram, there are interval cancers caused by reader error to minimal signs or complete masking when the lesion is obscured by dense breast tissue.  %commonly extensive mammogram breast density reside.

\begin{figure}[t!]
\centering
\scriptsize
%\begin{adjustbox}{width=1.01\textwidth}

\begin{tabular}{@{}c@{\hspace{1mm}}c@{\hspace{.4mm}}c@{\hspace{4mm}}c@{}}
%\rowfont{\normalsize}%
%&
&
Large invasive cancer &
Same cancer, likely masked &
Tumor size \& frequency \\
%Prior screening image & \\
% & cancer & missed by radiologist & \\
%$\rotatebox{90}{\hspace{4mm}} 
%&
\includegraphics[height=37mm]{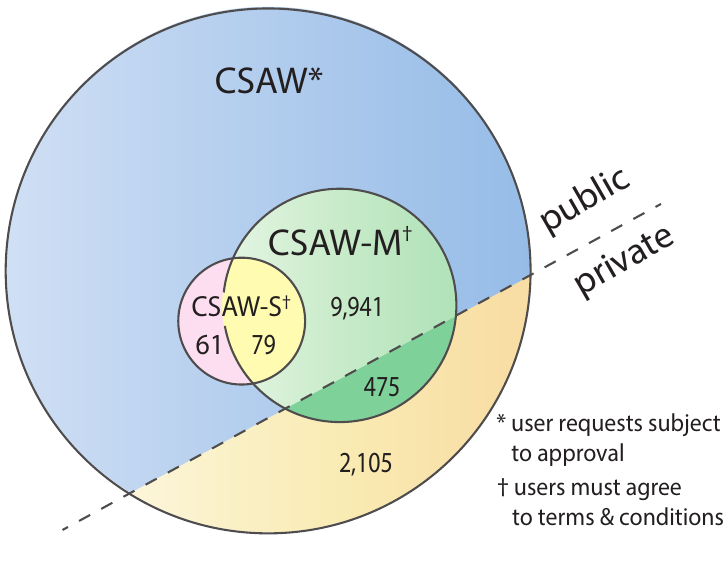} & 
\includegraphics[height=38mm]{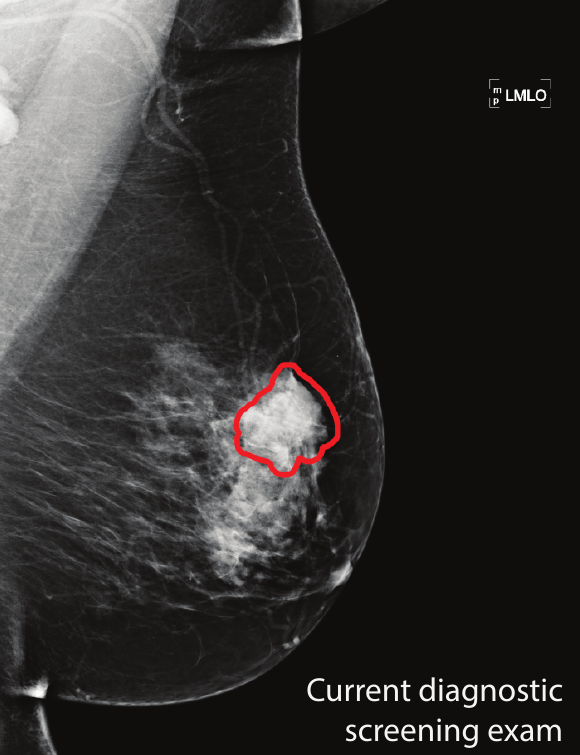} &
\includegraphics[height=38mm]{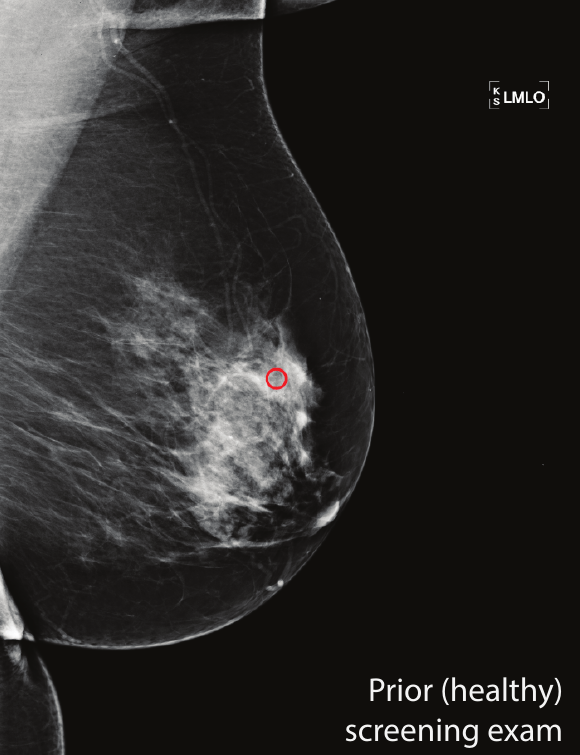} &
\includegraphics[height=38mm]{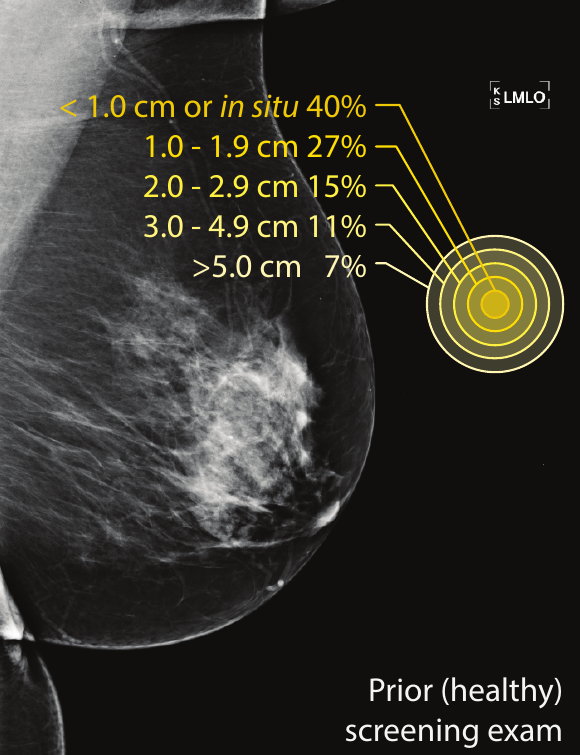} \\
\end{tabular}
%\end{adjustbox}
\normalsize

\caption{\emph{(left)} The CSAW family of datasets with number of screening participants indicated. \emph{(right)} From left-to-right, a mammogram of a large invasive cancer, the prior exam of the same cancer where the tumor was likely masked, and a visualization of tumor size \& frequency at detection~\cite{welch2016breast}. Note that, to maintain a pure notion of masking potential, CSAW-M images do not depict any identified cancers.}
\label{fig:sketch}
\vspace{-3mm}
\end{figure}

Masking refers to the phenomenon in which a tumor is hidden by the surrounding breast tissue, causing the cancer to be difficult or even impossible to discern with regular mammography, as seen in Figure~\ref{fig:sketch}. 
Masking can also result in \emph{large invasive cancers}\footnote{We define large invasive cancers as those confirmed to have spread and be $\geq$ 2cm at time of diagnosis.} -- a small cancer may be difficult to discern in certain images, allowing it to grow to a more lethal size.
%also represent a failure of the screening process -- the aim is to catch cancer early when it is easier to treat. 
%which hence makes cancer indiscernible during regular mammography. 
Masking is correlated with breast density,
as it has been shown that cancer in dense breasts is more likely to be missed during screening~\cite{holland2017quantification, destounis2017using, alonzo2019investigating}.
Density can be subjectively assessed by radiologists via the \emph{BI-RADS density} standard (ACR)~\cite{d2003breast, sickles2013acr}, or measured by automated tools such as Libra \cite{keller2015preliminary}. 
%BI-RADS density is widely used in the US and in some countries in Europe~\cite{d2003breast, sickles2013acr}. 
%Alternatively, breast density can be computed using tools such as Libra \cite{keller2015preliminary}.
These density measurements, however, do not perfectly correlate with masking potential.
Radiologists consider the distribution and pattern of tissue %in addition to density 
when assessing masking potential,
and have called for automated methods to assess the masking effect~\cite{conant2018beyond}.
%, while BI-RADS only considers the ratio of fibroglandular tissue to fat \ks{[cite? ask Fredrik]}.
Until now, the question of exactly how masking potential should be quantified remains an open one,
%there does not exist a direct measurement of masking potential, 
although some subjective notion has been added to certain categories of the most recent edition of BI-RADS density~\cite{spak2017bi}.

The ability to assess masking potential is crucial because it can identify screening participants most likely to benefit from supplementary radiological methods, \eg MRI. 
MRI is more sensitive than mammography, and has been proven to reveal tumors missed in regular mammographic screens~\cite{gordon2020mri}.
Unfortunately, MRI is too costly and cumbersome to screen the whole population.
The ability to predict high masking potential would allow clinics to identify screening participants most likely to benefit from MRI screening.
These participants could be offered additional screening, potentially detecting more cancers as demonstrated in the DENSE trial~\cite{gordon2020mri}.
Additionally, the ability to identify screening participants with low-masking potential -- fatty breasts where tumors are obvious -- would help hospitals more effectively allocate  radiological expertise.

\begin{figure}[t!]
%\footnotesize
% \normalsize
\centering
\begin{adjustbox}{width=1.01\textwidth}
\hspace{-3mm}
\begin{tabular}{@{}c@{\hspace{.4mm}}c@{\hspace{.4mm}}c@{\hspace{.4mm}}c@{\hspace{.4mm}}c@{\hspace{.4mm}}c@{\hspace{.4mm}}c@{\hspace{.4mm}}c@{}}
%\rowfont{\normalsize}%
%&
Level 1 &
Level 2 &
Level 3 &
Level 4 &
Level 5 &
Level 6 &
Level 7 &
Level 8 \\
%$\rotatebox{90}{\hspace{4mm}} 
%&
\includegraphics[height=50mm]{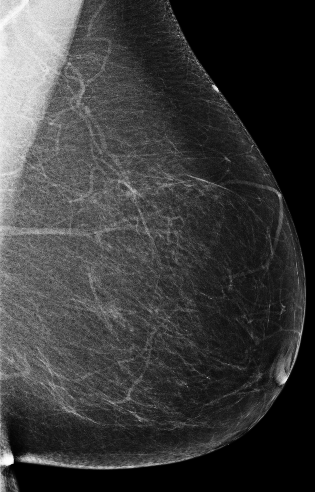}&
\includegraphics[height=50mm]{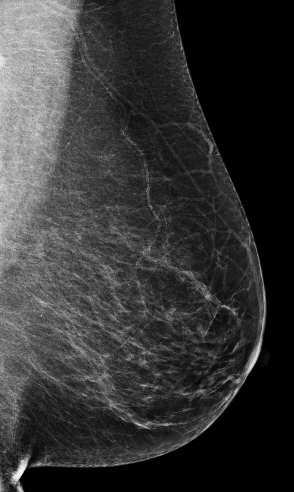}&
\includegraphics[height=50mm]{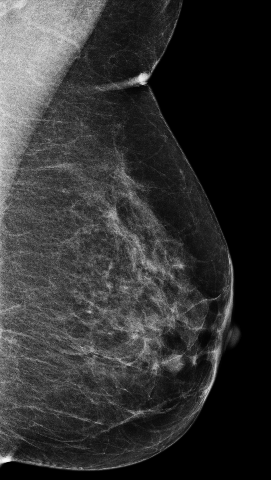}&
\includegraphics[height=50mm]{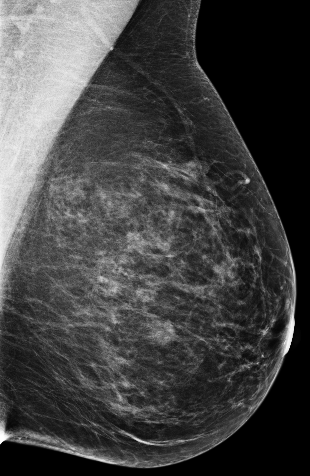}&
% \\
% &
% \\
% \rotatebox{90}{\hspace{3mm} } &
\includegraphics[height=50mm]{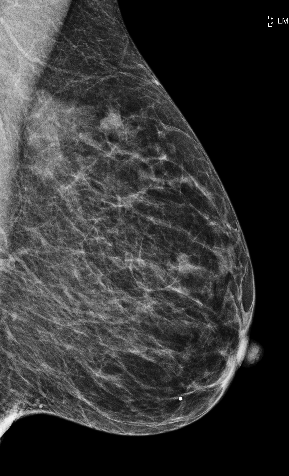}&
\includegraphics[height=50mm]{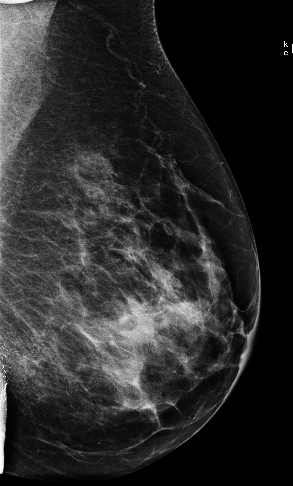}&
\includegraphics[height=50mm]{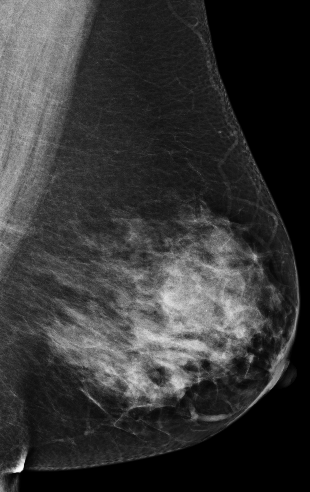}&
\includegraphics[height=50mm]{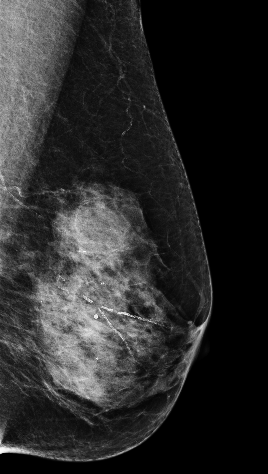}

\end{tabular}
\end{adjustbox}
\caption{Masking potential, or the possibility for cancer to be obscured in a mammogram, from CSAW-M. From left to right, the most agreed-upon images among five expert annotators for level 1 (lowest masking potential, easiest to assess) to level 8 (highest masking potential, hardest to assess).}
%Most agreed-on images for each masking level on CSAW-M. This figure shows, from left to right, level 1 (lowest masking level, easiest to asses) to level 8 (highest masking level, hardest to assess).  }
\label{fig:main_figure} 
\vspace{-4mm}
\end{figure}

% Several approaches to measure breast density are available. 
% One method is to apply automated algorithms to calculate breast density using tools such as Libra \cite{keller2015preliminary}.
% BI-RADS density, a four-scale visual assessment performed by radiologists, is widely used in clinical practice \cite{d2003breast, sickles2013acr}. 
% Masking potential is considered in some of the categories in the most recent edition of BI-RADS density \cite{spak2017bi}.
% A handful of datasets such as CBIS-DDSM \cite{lee2017curated}, INbreast \cite{moreira2012inbreast} and MIAS \cite{suckling1994mammographic} include breast density scores as metadata, but their limited data scale makes them difficult to use for developing deep learning models. 

%To study masking risk factors, several studies have shown that women with dense breasts are more exposed to the risk of interval cancers \cite{holland2017quantification, destounis2017using, alonzo2019investigating}, and combining breast density with other risk factors such as age and body mass index showed improved performance in detecting interval cancers \cite{mainprize2019prediction}. 
%Nevertheless, how the masking effect should be measured directly remains an open question. 

In this work, we introduce the \textbf{CSAW-M dataset} -- a collection of over 10,000 mammographic images and associated masking assessments from experts.
The assessments were graded by radiologists according to 8 levels of masking potential, as depicted in Figure \ref{fig:main_figure}, from easily assessed mammograms with low-masking potential (level 1) to difficult-to-assess examples with high-masking potential (level 8). 
This data can be used to train models capable of predicting masking potential from mammographic images in an ordinal classification setting.
%Furthermore, we provide associated clinical endpoints, including whether or not the women in the dataset developed interval or large invasive cancers.

% There is a lack of mammography datasets that are useful for benchmarking masking effect. To help address this issue, we introduce our dataset CSAW-M - a large-scale ordinal classification dataset for benchmarking mammographic masking of cancer. 
% \ks{We don't model masking of cancer, but masking potential. We need to make this clear.}
% In this dataset, we sort images into 8 levels of masking (Figure \ref{fig:main_figure}), where images in the first level are considered the easiest and the ones in the last level are the hardest to assess by human experts. 
% %To form our dataset, each time we gave radiologists mammogram pairs and asked them to mark the mammogram that is harder to assess. 
% Similar to BI-RADS, CSAW-M consists of subjective visual assessment by radiologists.
% However, BI-RADS mainly considers the amount of fibroglandular tissue relative to fat, while our dataset is a direct measure of potential masking. 

%The contributions of our work are as follows:
The unique features of CSAW-M include:
\begin{enumerate}
    \item It is the first dataset to directly address masking potential in mammography using expert assessments.
    % \item We offer annotations from five radiologists for our test set of 499 images, making this dataset useful to study annotator bias or agreement. 
    \item Aside from the masking potential assessments from five experts, CSAW-M also includes objective \emph{clinical endpoints}, \ie data on whether the screening participants developed interval or large invasive cancers.
    % \item Our dataset is large-scale, real-world and can be used for pre-training models for tasks such as cancer detection and breast cancer risk estimation. 
    %\item We introduce baseline implementations and evaluation of deep learning models to categorize images into different levels of masking.
    \item CSAW-M is the largest public collection of mammograms, containing digital mammograms from over 10,000 screening participants, which can be repurposed for other tasks.
    %It can be repurposed, \eg to improve self-supervised pretraining for a cancer detector, as described below.
    \item CSAW-M is distributed with a public test set for researchers to benchmark themselves. In addition, we defined a private test set which will not be distributed. 
    An evaluation service hosted at SciLifeLab will allow researchers to submit Dockers containing their models for evaluation on the private test set, as a control against overfitting to the public test set.
    % it will allow researchers to benchmark their models in a controlled, less biased setting via a planned  service hosted at SciLifeLab.
    \end{enumerate}
In addition to these features, we provide a detailed analysis of expert agreement w.r.t.~masking potential, and compare their performance to a baseline model we developed.
We release the source code for our annotation tool, the implementation of baseline models and metric calculations, as well as the trained models\footnote{Code available at: {\maskingRepo}.}.
Furthermore, the data contained in CSAW-M is relevant to the following research areas:

%(1) To the best of our knowledge, this is the first dataset that uses valuable experts annotations to directly addresses the potential for mammography masking.
%(2) We offer annotations from five radiologists, making this dataset useful to study annotator bias or agreement. 
%(3) Aside from subjective experts assessment, we also include objective clinical endpoints i.e. interval cancers and large invasive cancers as metadata, 
%(4) Our dataset is large-scale, real-world and can be used for pretraining for tasks such as cancer detection and breast cancer risk estimation. 
%(5) We introduce baseline implementations and evaluation of deep learning models to categorize images into different levels of masking.

% \item Masking levels are ordinal, making this dataset useful 
% dataset with ordinal labels, can be used for developing ordinal classification networks 

% \yl{what important questions do we want to ask in introduction?}

% \yl{one question that might be interesting to ask is - is masking correlated with interval and/or large invasive cancers? we can raise the question first and by showing our results, we say they are correlated }

\begin{itemize}
    \item \textbf{Ordinal classification/point-wise ranking}: labels in CSAW-M are ordinally related (ordered from 1 to 8). 
    Few image datasets support the development and benchmarking of ordinal classification models, which is of value to the ML community.
    \item \textbf{Better pre-training}: recent works have shown that ImageNet \cite{deng2009imagenet} pre-training can be outperformed by pre-training on datasets of similar domain, which provides a better initialization \cite{sowrirajan2021moco}. 
    CSAW-M can be valuable for pre-training models for tasks in a similar domain, \eg cancer detection in mammograms.
    \item \textbf{Noisy labels and annotator agreement}: the masking potential labels in CSAW-M are subjective assessments.
    As such, the opinions from 5 expert radiologists we collected can be valuable for researchers investigating the effects of human noise and bias in the annotation process (and ways to mitigate these effects or use them for modelling aleatoric uncertainty).
    % makes training a model happen under real-world noisy labels.
    % \item \textbf{Study annotator agreement and bias}: 499 test images in our dataset are annotated by 5 expert radiologists, allowing studying annotator agreement and bias.
%    \item \textbf{Clinical studies}: CSAW-M includes clinical endpoint metadata (whether women develop \textit{interval} or \textit{large invasive} cancers \footnote{We define large invasive cancers as those confirmed to have spread and be $\geq$ 2cm at time of diagnosis.}) which can prove valuable for clinical studies.
%    we include clinical endpoints such as whether an image is from a cancer patient and more specifically, whether a patient with interval and/or large invasive cancer, making this dataset valuable for clinical studies.
\end{itemize}

%\footnote{The dataset can be downloaded at \url{https://www.scilifelab.se/data/repository/}. Annotation tool and baseline implementations are also available at \url{https://github.com/} .\ms{we will not put any link to dataset in the paper during review}}.

%\ks{Add a sentence or two here about how CSAW-M can be accessed!!!}
CSAW-M is publicly available for non-commercial use \footnote{Dataset could be found at: {\datasetDOI}}.
It is hosted by the SciLifeLab Data Repository, a Swedish national infrastructure for sharing life science data.
% The site contains instructions on how to request the data.
% The data is self-contained -- enabling users to easily understand the content and organization of the files using the provided metadata file.
A \emph{datasheet}~\cite{gebru2018datasheets} summarizing CSAW-M, along with detailed documentation, can be found in the Appendix~\ref{datasheet}.
%\ref{datasheet}.

%with instructions on how to request the data \ms{(if dataset is not published by Monday night, we should say that we will provide the public link later)}. The data is self-contained - enabling users to easily understand the content and organization of the files using the provided metadata file. Also, detailed documentation of the dataset could be found in the Appendix.

Although CSAW-M represents the largest public collection of mammographic images, a number of other mammography datasets exist.
These datasets, summarized in Table~\ref{tab:datasets}, vary by modality, number of patients, demographics, and metadata provided.
Most are focused on tumor detection, although some provide density measures along with the metadata.
Existing public datasets are limited by the number of examples, the modality (scanned film is of inferior quality to digital mammograms), and the lack of explicit masking assessments. 
It is important to note that, unlike other datasets, CSAW-M does not contain images of cancers, so it is not directly useful for cancer detection.
%\ms{Explicitly say this is not a cancer dataset...}

%However, the low number of women and failure to explicitly capture masking potential\footnote{BI-RADS density assessments collected before 2017 do not consider masking.} limit their usefulness.

%\ks{Add a section about other datasets, how much data they have, are they film, etc. ALSO MAKE A TABLE SUMMARIZING THEM.}
%A handful of datasets such as CBIS-DDSM \cite{lee2017curated}, INbreast \cite{moreira2012inbreast} and MIAS \cite{suckling1994mammographic} include breast density scores as metadata, but their limited data scale makes them difficult to use for developing deep learning models. \ks{they don't explicitly model masking}

%\ks{UNDER CONSTRUCTION BELOW THIS LINE -------}

%Similar to BI-RADS, CSAW-M consists of subjective visual assessment by radiologists.
%However, BI-RADS mainly considers the amount of fibroglandular tissue relative to fat, while our dataset is a direct measure of potential masking. 

\vspace{-3mm}
\setlength{\extrarowheight}{0.1cm}
\begin{table}[t]
\caption{Public mammography datasets.}
\label{tab:datasets}
\vspace{-4mm}
\begin{center}
\scriptsize
\begin{tabular}{@{}l@{\hspace{1mm}}l@{\hspace{2mm}}l@{\hspace{1mm}}l@{\hspace{1mm}}l@{\hspace{1mm}}l@{\hspace{1mm}}l@{\hspace{1mm}}l@{}}
\toprule
& Year & Origin & Participants & Images & Image modality & Masking metadata & Cancer metadata \\
\midrule
MIAS \cite{suckling1994mammographic} & 1994 & UK & 161 &  322 & Film & Non-ACR density & Cancer center \& radius\\
LLNL \cite{oliveira2008toward} & 1995 & USA & 50 &  198 & Film & Non-ACR density \& subtlety \footnotemark & Cancer ROI, pixel-level\\

BancoWeb 
\cite{matheus2011online} & 2010 & Brazil & 320 &  1,473 & Film & Non-ACR density & ROI available in a few images\\
INBreast \cite{moreira2012inbreast} & 2012 & Portugal & 115 & 410 & Digital & ACR density & Cancer ROI, pixel-level \\
BCDR \cite{lopez2012bcdr} & 2012 & Portugal & 1,734 & 7,315 & Film \& Digital  & ACR density & Cancer ROI, pixel-level \\
CBIS-DDSM \cite{lee2017curated} & 2017 & USA & 1,566 &3,103 & Film & ACR density \& subtlety  & Cancer ROI,  pixel-level\\
CSAW-S \cite{matsoukas2020adding} & 2020 & Sweden & 140 &274 & Digital & - & Cancer ROI, pixel-level \\
CSAW-M & 2021 & Sweden & 10,020 &10,020 & Digital & Explicit expert assessment & Interval/large cancers, image-level \\
\bottomrule
\normalsize
\end{tabular}
\end{center}
\vspace{-5mm}
\end{table}
\footnotetext{Subtlety is a subjective rating of difficulty in viewing the abnormality in the image, as defined in \cite{oliveira2008toward, lee2017curated}, while masking potential discussed in this paper considers cancer-free mammograms.}

\newpage
\section{CSAW-M dataset creation}\label{sec:datacreation}

\begin{figure}
    \centering
    \hspace{-1mm}\includegraphics[width=1.02\linewidth]{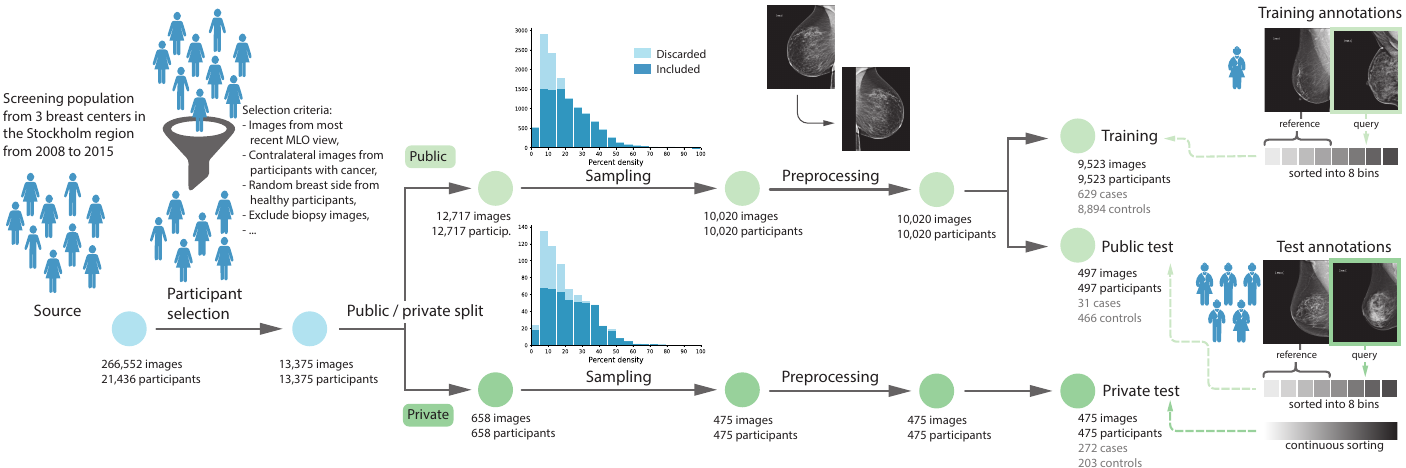}
    \caption{Overview of the creation of CSAW-M. 
    From a screening population in Stockholm, participants were selected based on criteria described in Appendix \ref{sec:image_selection_procedure}.
    The data was divided into public and private sets and then sampled to obtain a more uniform distribution of breast density.
    The images from the selected participants were preprocessed and annotated by experts -- sorted using pairwise comparisons into 8 bins (public) or fully sorted (private).
    Please refer to the text for details.}
    \label{fig:before_sampling_hist}
    \vspace{-3mm}
\end{figure}

%\subsection{CSAW dataset}
CSAW-M consists of screening mammograms along with metadata describing expert masking potential assessments, clinical endpoints, density measures, and image acquisition parameters.
%\yl{(just to double check- we don't need to mention libra densities that we included in the csvs here right?)}.
CSAW-M is part of an ecosystem of mammography datasets based on the CSAW population-based cohort~\cite{dembrower2019multi}, depicted in Figure~\ref{fig:sketch}.
CSAW is a collection of millions of screening mammograms of screening participants aged 40 to 74 gathered from the three breast centers of the Stockholm region between 2008 and 2015.
The CSAW \emph{case-control} dataset, hereafter referred to as CSAW for brevity, is a subset of the full CSAW cohort containing all cancers, along with a random sampling of healthy screens from the Karolinska breast center.
A portion of CSAW (2,580 screening participants) is designated as a private held-out test set, unavailable to the public, for controlled benchmarking of various tasks.
CSAW-M is subset of CSAW, created here, to study masking.
It is divided into a training set (9,523 examples), a public test set (497 examples), and a private test set corresponding to those in CSAW (475 examples).
A summary of the dataset is provided in Table~\ref{tab:dataset_summary}.
CSAW-M partially overlaps with 
CSAW-S, a sister dataset focused on segmentation in mammograms~\cite{matsoukas2020adding}.
Below, we describe the procedure followed to create CSAW-M depicted in Figure \ref{fig:before_sampling_hist}.

\vspace{-3mm}
\paragraph{Image selection.}
Screening participants from CSAW were selected for inclusion to CSAW-M according to a flowchart found in Appendix~\ref{sec:image_selection_procedure}.
%Recall that the women in the CSAW-M private test set correspond to the \textit{private} split of the CSAW dataset, while the public train and test sets relate to the \textit{non-hidden} split of CSAW.
As shown in Figure \ref{fig:before_sampling_hist}, starting from the CSAW population, we selected participants with mammographic screening exams from Karolinska University Hospital acquired with Hologic devices after the data was curated. 
%We considered women in the private split of CSAW to form our private test set and women in its public portion to form our public train and test sets. 
From these sets of participants, we selected images as follows:
the most recent mediolateral oblique~(MLO) view of the breast was included, since MLO offers the best visualization of the breast~\cite{eklund2000art}.
If a selected participant had cancer, we selected the image of the contralateral breast~(the one without cancer) to avoid contaminating the masking potential annotation task with actual tumors. Otherwise, the image was chosen with a random breast side. 
This resulted in screening participants fitting our selection criteria. To form our private test set, we finally sampled from the participants who correspond to the private split. From the participants belonging to the public split, we included all with cancer and sampled from the healthy population (as described below) to form our public data.

%From these women, we sampled from the private split of CSAW to form the CSAW-M private test set.
%For the public data, we included all women with cancer and sampled from the healthy population.

%First, women with mammographic screening exams from Karolinska University Hospital acquired with Hologic devices were selected. From the selected women, we uniformly sampled to form our private test set. For the public data, however, all women with cancer diagnoses were included, and only healthy women were sampled. From the sampled women, the most recent mediolateral oblique~(MLO) view of the breast was included, since MLO offers the best visualization of the breast~\cite{eklund2000art}.
%If a selected patient had cancer at their most recent screening, we selected the contralateral breast (the one without cancer) from a prior exam to avoid contaminating the masking potential annotation task with actual tumors. \yl{to do - motivate why contralateral is a good approximation, - citation if possible \cite{byng1996symmetry}}
%Otherwise, the breast side was randomly chosen.

The images selected with our selection criteria above had a strong positive skew (light blue in Figure \ref{fig:before_sampling_hist} and Figure \ref{fig:sampling_dists} of the Appendix) 
in terms of percent breast density computed by Libra~\cite{keller2015preliminary}.
The most clinically interesting samples -- very dense and very fatty breasts -- belong to the under-represented tails of the distribution. We under-sampled the center of the distribution while keeping all samples of the tails, which resulted into a more uniform distribution (dark blue). Figure \ref{fig:before_sampling_hist} shows an overview of the selection procedure.
%See Appendix~\ref{sec:sampling_and_preprocessing} for details.

\vspace{-3mm}
\paragraph{Image preprocessing.}
%The selected images were processed according to a procedure described in Appendix~\ref{sec:sampling_and_preprocessing}.
%This including privacy measures such as downsampling and removal of identifiers.
The source images of our dataset are DICOM format files which are resized to $632\times512$ and saved with 16-bit PNG format as raw data of CSAW-M. Using the DICOM metadata, we perform a horizontal flip to make all breasts left-posed and rescale the intensity linearly into a proper DICOM window range. 
We locate the centroid of the breast and move it horizontally to the center of the image. Zero-padding is applied on the images in order to ensure uniform size among the images.
The text in images (which includes the initials of the technician, breast laterality and view position) is removed by extracting the contour that is the closest to the top right corner of the image. Finally, the preprocessed images are saved as 8-bit PNGs. Further details are provided in Section \ref{sec:sampling_and_preprocessing}
 of the Appendix.

\begin{table}[t]
\caption{Summary of the CSAW-M dataset.}
\label{tab:dataset_summary}
\vspace{-4mm}
\begin{center}
\scriptsize
\begin{tabular}{@{}l@{\hspace{2mm}}c@{\hspace{2mm}}c@{\hspace{2mm}}c@{\hspace{2mm}}c@{\hspace{2mm}}c@{\hspace{2mm}}c@{\hspace{2mm}}c@{\hspace{2mm}}c@{\hspace{2mm}}c@{}}
\toprule
%& \rot{Ref.} & Year & Origin & Women & Image modality & Masking metadata & Cancer metadata \\
& \# images & Resolution  & \# interval / large & \# composite & \# & \#  masking & Masking & Metadata & Publicly \\
&  &  & invasive / total cancers & endpoints & controls & annotations & levels & & available? \\
\midrule
Public train & 9,523 & 632$\times$512 & 148 / 279 / 629 & 347 & 8,894 & 1 / image & 1-8 & Density, acquisition & Yes \\
Public test & 497 & 632$\times$512  & 11 / 13 / 31 & 19 & 466 & 5 / image & 1-8 & Density, acquisition & Yes \\
%Total & 10,020 & 632$\times$512 & PNG & 159 & 292 & 366 & 9,360 & mixed & Density, image & Yes \\
\midrule
Private test & 475 & 632$\times$512  & 81 / 111 / 272 & 158 & 203 & 5 / image & 1-475 & Density, acquisition & No \\
\bottomrule
\normalsize
\end{tabular}
\end{center}
\vspace{-5mm}
    
\end{table}

%\subsection{Annotation procedure}
\vspace{-3mm}
\paragraph{Annotation procedure.}
%We detail the annotation procedure for the private test set below, followed by the public sets.
%In this part, we detail the annotation procedure for the CSAW-M dataset. We will first describe how the private test set (hold-out set) was annotated, and then we will outline the annotation procedure for the public test and training set. 
The goal of the annotation procedure was to label each image with expert assessments of masking potential.
Masking was quantized into 8 bins, or \emph{levels}, as depicted in Figure~\ref{fig:main_figure}, for the public training and test sets.
Images in the private test set are fully sorted according to masking. 
%\ks{then quantized post-hoc} \ms{is it not better just to talk about private test ranks than bins?}.
Individually sorted examples provide a more granular assessment, but at the cost of increased annotation time.
We opted for fine granularity on the private test set because \emph{(1)} it allows for a more fine-grained assessment, and \emph{(2)} it allowed us to identify robust initial bins\footnote{We use ``bin'' and ``level'' interchangeably to denote collections of images with similar masking potential.} for the 8 masking levels in the public training/test sets.
To represent the initial bins, we chose images personalized to each radiologist, but with highest agreement among the other experts.
The benefit to this approach is that the starting point respects subjective assessments while at the same time choosing representative examples for each masking level.

%The benefit of having personalized bins as a starting point is that they are constructed based on \textit{individual} sorting, so it respects the subjective assessments. 
%At the same time, choosing the most \textit{agreed-on} images ensures the initial bins have better representative mammograms for each masking level.

%Importantly, these sorted images provide a fine-grained subjective assessment which we adapted to provide each radiologist with robust, personalized initial bins that \emph{agree} with the other experts for labeling the public sets.

%into similar bins as the other experts.

%\ms{The benefit of having personalized bins as a starting point is that they are constructed based on \textit{individual} sorting of the private test set (so it respects the subjective assessments). At the same time, choosing the most \textit{agreed-on} images as the starting point would allow the initial bins to have better representative mammograms of the corresponding masking level.}

Five experts contributed annotations to CSAW-M, each a licensed radiology specialist from the Stockholm region.
Their experience in breast radiology ranged from 2 years to over 30 years.
Note that the annotation of masking potential is a novel task for the radiologists. That is, it is not performed in their routine clinical practice, nor was it a part of their training.
The training set contains one expert annotation per image, while the public and private test sets have five per image.
We provide the individual assessments as well as a \emph{ground truth} for the test set, computed as the median annotation. 
The median was chosen because \emph{(1)} it is robust to outliers, \emph{(2)} if there is a majority vote, median always selects it, and \emph{(3)} it simplifies the process of discretizing masking levels.
%it simplifies the process of quantizing the score into discrete masking levels.

%\ms{(justifications of why median: 1) median is a measure that is robust against outliers (as opposed to mean) \ms{(maybe cite works)}, 2) if there is a majority in the votes, median will always select that, 3) taking median does not require additional steps to convert the final label back to a discrete level, while taking mean often requires rounding the number to a discrete label, and deciding the best way of rounding is not trivial.)}.
%\ks{How to handle the private test set?} \ms{I think we can simply say it's media of ranks}

The annotation procedure itself was based on a principle of \emph{pairwise comparisons}.
As depicted in Figure~\ref{fig:before_sampling_hist}, an annotation software tool presented radiologists with a pair of images, a query {\q} and a reference $r$  (see Appendix \ref{annotation_procedure_details} details). 
The radiologist was tasked with \emph{deciding which image has the higher masking potential} -- or, put another way -- \emph{which image is harder to be certain there is no tumor?} % \yl{Here?} .
Based on the experts response \emph{(query {\q}, reference $r$, or ``no difference")}, the query image was sorted relative to the reference image.
Through a series of such comparisons, similar to a binary or ternary search, images were sorted either into 8 masking levels, or down to the individual images, see Figure \ref{fig:ternary}.
We use pairwise comparisons because they are more meaningful and repeatable for the experts than the arbitrary assignment of ordinal labels.

We chose a granularity of 8 masking levels for several reasons.
Eight masking levels meant that, at most, 3 pairwise comparisons were necessary to sort each image. 
%\yl{(refer to Figure \ref{fig:before_sampling_hist} ?)}. 
This appeared to be an acceptable compromise between granularity and annotation cost, as higher granularity appeared to limit the tendency of the experts to agree. %push the limit of agreement among experts.

\vspace{-3mm}
\paragraph{Private test set.}
Creation of the private set started with a common \textit{seed list} of 6 sorted mammograms selected by one of the experts.
Each radiologist was given this list, and expanded it to a list of 500 individually sorted images through pairwise comparisons, via a strategy described below. 

Given a query image {\q}, we try to find a suitable position to place it so that the list remains sorted.
This is accomplished by comparing {\q} with multiple images from the list, as shown in Figure~\ref{fig:ternary} of Appendix \ref{annotation_procedure_details}.
To enforce consistency in the annotations, we devised a method inspired by the \emph{ternary search} algorithm.
Suppose we want to insert $q$ in the interval $\left[l, h\right]$.
We compare $q$ against two \textit{anchor} images $a_1$ and $a_2$ positioned at the $p_1 = \left(l + h\right)/3$ and $p_2 = 2 \cdot \left(l + h\right)/3$.
The ternary comparison amounts to two consecutive pairwise comparisons presented to the expert, where the query image $q$ is compared against each anchor, $a_1$ and $a_2$ consecutively.
The answers determine where $q$ is placed, and logical checks on the answers ensure the expert answered consistently.

Ternary comparisons ensure annotator consistency, but are costly. 
Moreover, as the depth of the search increases, expert self-consistency decreases.
Hence, we used ternary search for the first two steps of each sorting, after which we employed a \emph{binary search} based method.
In the binary search, given the search interval $\left[l, h\right]$, the image at position $p=\left(l+h\right)/2$ would be selected as the \textit{reference} image. %Figure \ref{fig:before_sampling_hist}(d) shows comparisons and outcomes, similar to Figure \ref{fig:before_sampling_hist}(c). 
Technical details of the ternary and binary search are provided in Appendix \ref{annotation_procedure_details}.

\vspace{-3mm}
\paragraph{Training and public test sets.}
Image-level sorting, as performed on the private test set, was too costly to apply to the 10,020 images set aside for public training and testing.
Therefore, for the public data, experts sorted images into discrete masking levels, with 8 levels chosen for the reasons described above.
To begin 
%annotating the public data into 8 masking levels, 
we first created a personalized list of 32 images, 4 per masking level, for each expert.
To accomplish this, each expert's sorted private list was divided into 8 equal bins.
The average rank was defined as: $\tilde{r}_i = \sum_{j=1}^{5}{r^j_i}/5$ where $r^j_i$ denotes the rank assigned by radiologist $j$ for image $i$ .
Similarly, the rank delta of an image $i$ w.r.t. different annotators was defined as: $\delta^j_i = |\tilde{r}_i - {r^j_i}|$.
Each expert received as a seed for annotating the public data, 32 images -- the top 4 of their own personally sorted images per bin, with lowest $\delta^j_i$ (the \textit{most agreed-on} images).

As before, our goal was to sort the images.
But this time, the search interval was over masking levels instead of fully sorted images.
Given a query image $q$ and 8 masking levels, the initial search interval would be [1-8].
For each step in the search, we first selected a \textit{reference bin} in the middle of the search interval, from which we took a random \textit{reference image}. 
The query image was then assessed against the reference image using a pairwise comparison.
Since we defined an even number of masking levels, a situation can occur where the middle of a binary search interval would lie between two bins.
Rounding would result in some bins being selected more often than others.
To prevent that, we make sure that bins have an equal chance of being selected.
For example, the initial search interval is [1-8], so we start by randomly selecting bin 4 or 5 as the reference bin.
Following that, whenever the middle of the binary search falls between two bins, we round it down if it is above 4 and round it up if it is below 5.
This simple modification allows for \textit{symmetrical} moving out from the middle and allows bins at different steps of the binary search to have an equal chance of being shown as reference.
See Appendix \ref{sec:experts_dist} for the masking level distributions of each expert resulting from the annotation process.

\vspace{-3mm}
\paragraph{Summary of the CSAW-M dataset.}
CSAW-M public data consists of 10,020 mammography images at 632$\times$512 resolution in 8-bit PNG format, and associated metadata as described in Table~\ref{tab:dataset_summary}.
The metadata 
%is provided as a CSV file, and 
includes the masking potential labels collected through the annotation process (including the computed ground truth); clinical endpoints \ie cancer attributes including \texttt{cancer}, \texttt{interval} and \texttt{large invasive}; image acquisition attributes including \texttt{laterality}, intensity window \texttt{center} and \texttt{width}; and density attributes including \texttt{percent density}, and \texttt{dense area}.
The annotations were equally distributed among the annotators, with approximately 2,500 assigned to each.
A few images included in the annotation process are not included in the final dataset for various reasons, \eg missing/declined annotations or problems with the image.
In general, the experts were able to agree as to the masking potential of mammography images, although some tended to agree more closely than others, as indicated in Figure \ref{fig:network performance}.
See the discussion in Section~\ref{sec:results}.

\section{Experiments}\label{sec:experiments}
% \begin{table}
%   \caption{Kendall correlation tau-b score on test set}
%   \label{kendall-table}
%   \centering
%   \includegraphics[height=2in]{figs/experiments/kendall.png}
% \end{table}
% \begin{table}[t]
% \centering
% \caption{F1 scores on extreme levels - private test}\label{table:f1}
% \scriptsize

% \begin{tabular}{cccc}
% \toprule
%                          &           & F1 on level 1-2 & F1 on level 7-8 \\ \midrule
% \multirow{5}{*}{Experts} & Expert 1  & 0.8225          &  0.6400         \\
%                          & Expert 2  & \textbf{0.8485} & \textbf{0.7600}          \\
%                          & Expert 3  & 0.6926          & 0.4800          \\
%                          & Expert 4  & 0.7792          & 0.5500          \\
%                          & Expert 5  & 0.7532          & 0.6400         \\ \midrule
% \multirow{2}{*}{Models}  & Softmax   & 0.7489          & 0.6537          \\
%                          & Multi-hot & 0.7461          & 0.6329 \\
% \bottomrule
% \normalsize
% \end{tabular}
% \end{table}

We conducted experiments using simple models to \emph{(1)} empirically analyze CSAW-M, and \emph{(2)} serve as baselines for future work.
We evaluated performance of these models for prediction of masking potential, as well as for the indirect clinical tasks of correlating masking potential estimates with the two clinical endpoints of interval cancers and large invasive cancers.

\vspace{-3mm}
\paragraph{Baseline models.}
We trained two baseline models to predict masking potential from mammography images. Both use ResNet-34 \cite{he2016deep} as the backbone.
\vspace{-2mm}
\begin{itemize}
    \item \emph{ResNet34 one-hot} -- This model uses a standard approach for categorical classification, where each class is treated independently.
    Masking levels are predicted independently using standard softmax and cross-entropy loss to predict one-hot encodings.
    \item \emph{ResNet34 multi-hot} -- This model accounts for the ordinal relation between masking levels using \emph{multi-hot} encodings~\cite{cheng2008neural}.
    Training a model with multi-hot encoding could be seen as multi-label classification, where the task of classifying an item into $K$ ordinal classes is equivalent to solving $K-1$ independent binary classifications, each class being a superset of the previous.
    For an ordinal classification problem with $K$ classes, the multi-hot encoding for datapoint $x_i$ with ordinal label $l_i \in \{1, ..., K\}$ is defined as $\{y_i^1, y_i^2, ..., y_i^{K-1}\}$ where for each $ k \in \{1, ..., K-1\}$, $y_i^{k} = \mathbbm{1}\{l_i > k\}$ where $\mathbbm{1}\{.\}$ denotes the indicator function.
    %which evaluates to 1 if the condition is satisfied, and 0 otherwise. 
%     Multi-hot encoding has proven effective in other ordinal classification tasks such age estimation from face images \cite{niu2016ordinal}. Inspired by that, we also employ it in our ordinal classification setup.
\end{itemize}
\vspace{-2mm}
% \vspace{-3mm}
% \paragraph{Implementation details}
% 
\begin{figure}[t!]
%\footnotesize
% \normalsize
\centering
\begin{adjustbox}{width=1.01\textwidth}
\hspace{-3mm}
\begin{tabular}{@{}c@{\hspace{.4mm}}c@{\hspace{.4mm}}c@{\hspace{.4mm}}c@{\hspace{.4mm}}c@{\hspace{.4mm}}c@{\hspace{.4mm}}c@{\hspace{.4mm}}c@{}}
%\rowfont{\normalsize}%
%&
\textcolor{orange}{2} / 22322 / \textcolor{blue}{2} /  \textcolor{red}{2} &
\textcolor{orange}{4} / 44544 / \textcolor{blue}{4} / \textcolor{red}{4} &
\textcolor{orange}{8} / 88487 / \textcolor{blue}{8} / \textcolor{red}{8} &
\textcolor{orange}{4} / 41842 / \textcolor{blue}{2} / \textcolor{red}{3} &
\textcolor{orange}{8} / 83885 / \textcolor{blue}{5} / \textcolor{red}{5} &
\textcolor{orange}{5} / 65464 / \textcolor{blue}{8} / \textcolor{red}{7} &
\textcolor{orange}{7} / 66887 / \textcolor{blue}{5} / \textcolor{red}{5} &
\textcolor{orange}{7} / 77886 / \textcolor{blue}{4} / \textcolor{red}{5} \\

% \textcolor{orange}{2} / 22322 / \textcolor{blue}{2} /  \textcolor{red}{2} &
% \textcolor{orange}{4} / 41842 / \textcolor{blue}{2} / \textcolor{red}{3} &
% \textcolor{orange}{4} / 44544 / \textcolor{blue}{4} / \textcolor{red}{4} &
% \textcolor{orange}{5} / 65464 / \textcolor{blue}{8} / \textcolor{red}{7} &
% \textcolor{orange}{7} / 66887 / \textcolor{blue}{5} / \textcolor{red}{5} &
% \textcolor{orange}{7} / 77886 / \textcolor{blue}{4} / \textcolor{red}{5} &
% \textcolor{orange}{8} / 83885 / \textcolor{blue}{5} / \textcolor{red}{5} &
% \textcolor{orange}{8} / 88487 / \textcolor{blue}{8} / \textcolor{red}{8} \\

\includegraphics[height=50mm]{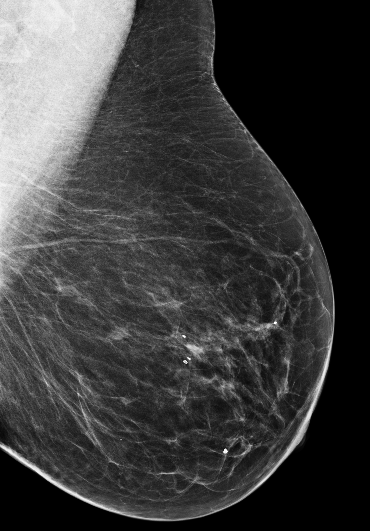}&
\includegraphics[height=50mm]{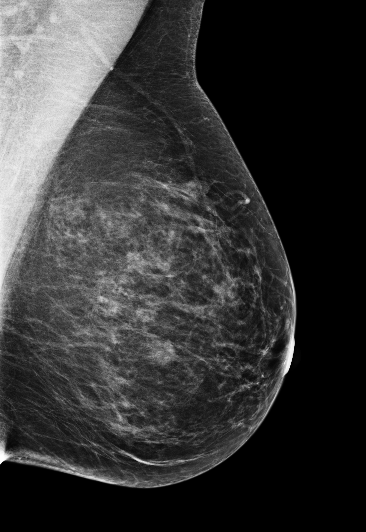}&
\includegraphics[height=50mm]{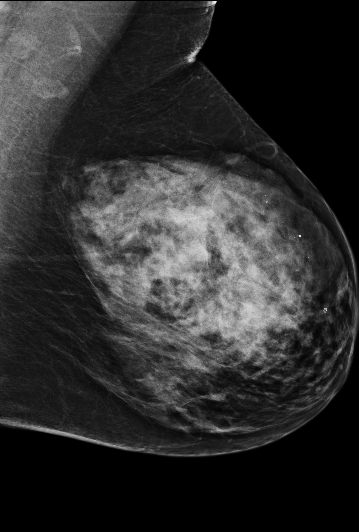}&
\includegraphics[height=50mm]{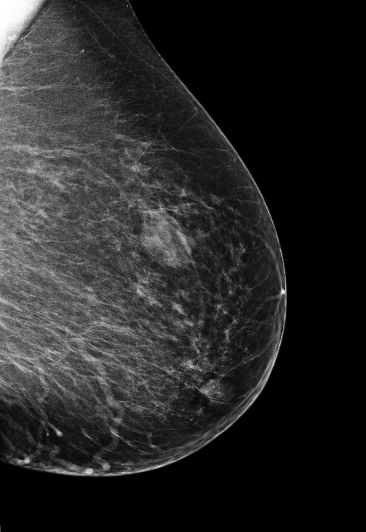}&
\includegraphics[height=50mm]{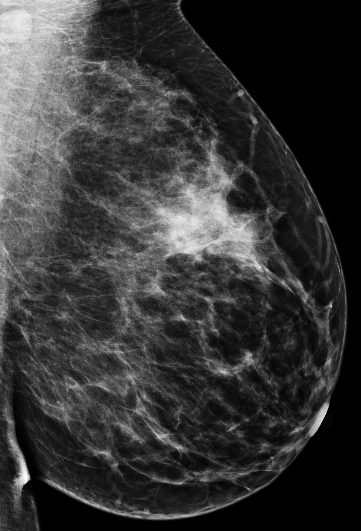}&
\includegraphics[height=50mm]{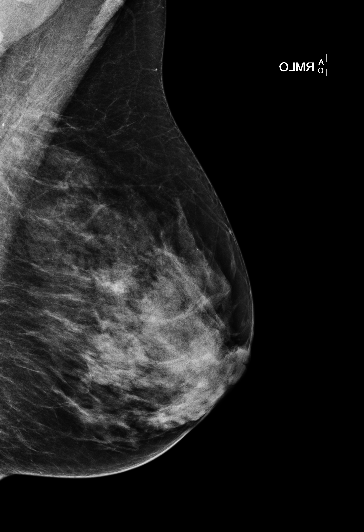}&
\includegraphics[height=50mm]{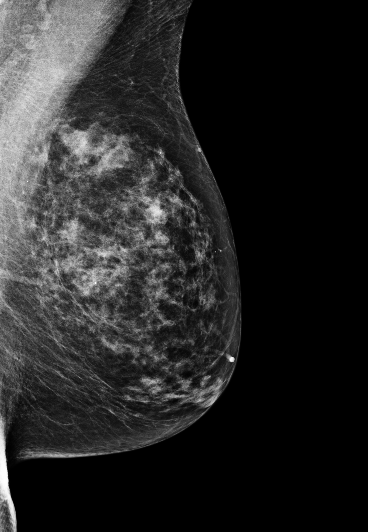}&
\includegraphics[height=50mm]{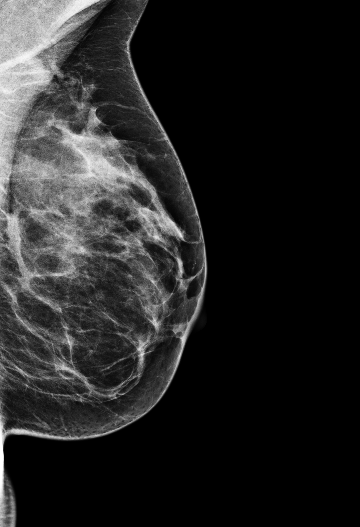}\\

\emph{(a)} Experts and models &
\emph{(b)} Experts and models &
\emph{(c)} Experts and models &
\emph{(d)} Outlier expert, &
\emph{(e)} Experts &
\emph{(f)} Models over- &
\emph{(g)} Models under- &
\emph{(h)} Models under-  \\
agree. &
agree. &
agree. &
possible error. &
disagree. &
estimate masking. &
estimate masking. &
estimate masking. \\

% \emph{(a)} Experts and models &
% \emph{(b)} Outlier expert, &
% \emph{(c)} Experts and models &
% \emph{(d)} Models over- &
% \emph{(e)} Models under- &
% \emph{(f)} Models under-  &
% \emph{(g)} Experts &
% \emph{(h)} Experts and models\\
% agree. &
% possible mistake. &
% agree. &
% estimate masking. &
% estimate masking. &
% estimate masking. &
% disagree. &
% agree. \\
\end{tabular}
\end{adjustbox}
\caption{Agreement and disagreement amongst experts and models. Eight mammograms are shown, along with labels as follows: \textcolor{orange}{GT} / experts 1-5 / \textcolor{blue}{one-hot} / \textcolor{red}{multi-hot}.}
%Most agreed-on images for each masking level on CSAW-M. This figure shows, from left to right, level 1 (lowest masking level, easiest to asses) to level 8 (highest masking level, hardest to assess).  }
\label{fig:discussion} 
\vspace{-4mm}
\end{figure}

For both models, we used a batch size of 64 and trained on 632$\times$512 images using the Adam optimizer \cite{kingma2014adam}. 
Both are initialized with ImageNet-pretrained weights \cite{deng2009imagenet}.
We used a learning rate of $1\mathrm{e}{-6}$ and applied random horizontal and vertical flipping, random rotation of 10 degrees, and small random brightness and contrast jittering as data augmentation. 
We used 5-fold cross-validation to determine the stopping iteration, which we used in the final training run using all the training data. 

\vspace{-3mm}
\paragraph{Task 1: Ordinal classification of masking potential.}
%\paragraph{Ordinal classification of masking potential}
The principal task of CSAW-M is to model the median expert opinion of masking potential level in the range [1-8].
Recall that the expert labels in CSAW-M are ordinally related, implying that a prediction confusing level 1 with level 8 is worse than one confusing level 1 with 2.
We consider two metrics widely used to evaluate ordinal classification, \emph{(1)} average mean absolute error (\textit{AMAE}) which measures the average distance of predicted classes w.r.t.~the true classes and is robust to class imbalance \cite{baccianella2009evaluation}, and \emph{(2)} {\kendall} \cite{kendall1938new} which measures the correlation of two rankings based on the number of concordant and discordant pairs.
{\kendall} ranges from -1 (perfect inverse correlation) to 1 (perfect correlation), and 0 indicates no correlation.
%\ks{Add a rule of thumb for {\kendall}?}

The metrics described above consider performance over all masking levels.
In addition, we consider model performance at identifying low- and high-masking mammograms.
From a clinical perspective, these levels are most interesting because they represent cases the experts are most and least confident about.
Participants with high-masking images can be \eg offered additional screening. 
We consider the two lowest masking levels (1 and 2) together as \textit{low-masking levels}, and the two highest ones (7 and 8) as \textit{high-masking levels}.
This choice was based on feedback from the experts.
To assess how the models perform at identifying images in these \emph{tail} masking levels, we use \emph{F1-score} -- a common metric to assess the performance in information retrieval.

\vspace{-3mm}
\paragraph{Task 2: Identification of interval and large invasive cancers.}
A secondary task of high clinical relevance is to measure the correlation between predicted masking estimates and certain cancers.
In particular, we consider how masking potential can predict \emph{interval cancers} and \emph{large invasive cancers} without being explicitly trained for these tasks.
We measure performance using area under the ROC curve (AUC) for individual cancer types and for the \emph{composite endpoint} (CEP) containing both types.
We also calculate the \emph{odds ratio} (OR), a measurement often used in clinical studies. 
The odds ratio is simply the ratio of the odds of an event occurring in one group to the odds of it occurring in another group.
To compute it, model predictions are divided into groups $g_i$. 
Here, groups correspond to quartiles of the model predictions, so $i \in \{1, 2, 3, 4\}$. 
For screening participants in group $g_i$, the odds of having interval cancer is computed as $O_i ={IC}_i / \tilde{{IC}_i}$, where ${IC}_i$ is the number of participants from $g_i$ with interval cancer, and $\tilde{{IC}_i}$ is the number without.
The odds ratio is then computed relative to a reference group, in this case the reference is the first quartile $g_1$, as $OR_i = O_i/O_1$.
Note that $OR_1 = 1$ for all models.
If a masking estimate is a  good predictor of interval cancer it will show strong odds ratios in the top quartiles and exhibit monotonically increasing odds ratios.

%exhibit monotonically increasing odds ratios, with strong odds ratios in the highest quartiles.

%If a masking estimate is a  good predictor of interval cancer it will exhibit monotonically increasing odds ratios, with strong odds ratios in the highest quartiles.

We compare our baseline models with \emph{dense area} and \emph{percent density} computed using Libra \cite{keller2015preliminary}.
This was done because density is known to be correlated with the clinical endpoints.
For a fair comparison, it was necessary to convert the discrete masking predictions from our models to continuous values.
We compute a continuous score as the weighted average of probabilities that an input belongs to each masking level.
Refer to Appendix \ref{details_on_models} for details.

\section{Results and discussion}\label{sec:results}

\begin{figure}[t]
\centering
\subfloat[Kendall's $\mathit{\tau_b}$ (higher is better) \label{fig:kendall}]{\includegraphics[width=0.4\columnwidth]{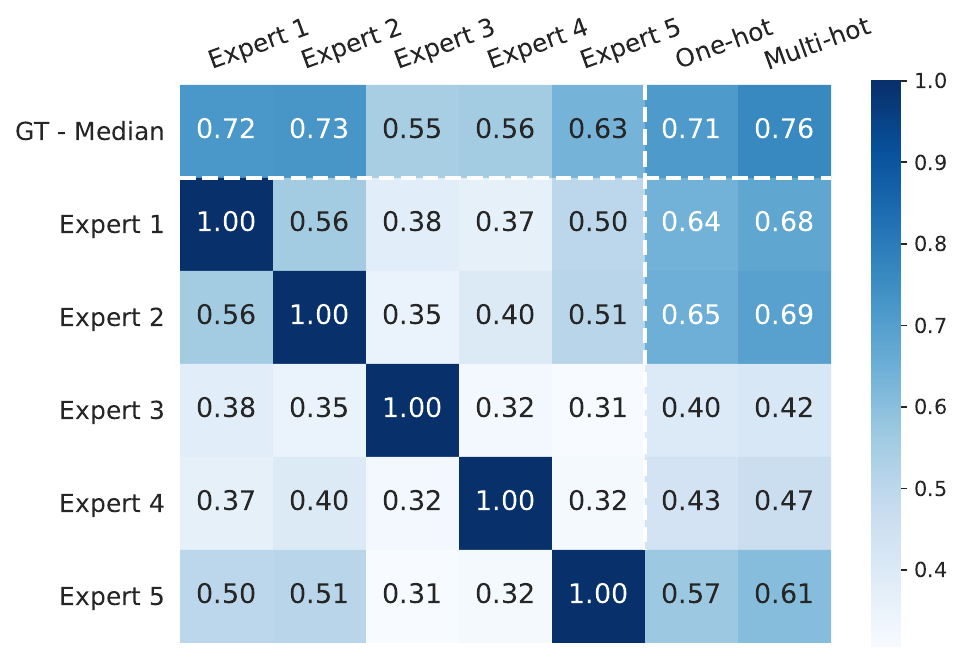}}\quad
\subfloat[{AMAE} (lower is better) \label{fig:amae}]{\includegraphics[width=0.4\columnwidth]{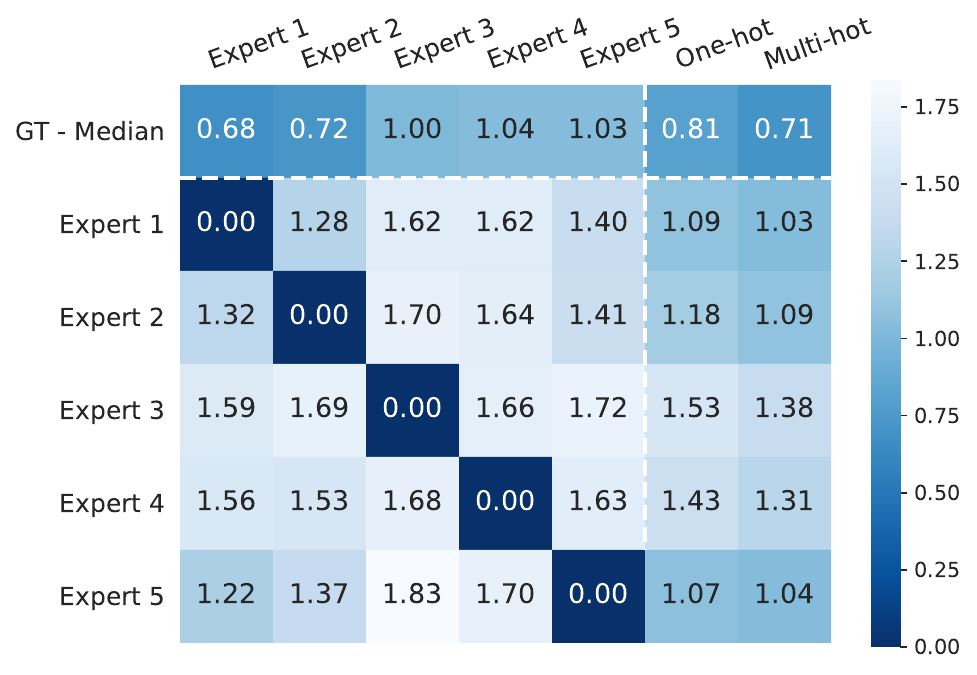}}\quad
\caption{Expert and model agreement on public test set. We perform five runs on our one-hot and multi-hot models and report the mean. See Appendix~\ref{sec:agreement_private} for results on the private test set.} 
\label{fig:network performance}
\vspace{-3mm}
\end{figure}

% \yl{what about p values? show in appendix?}
\vspace{-3mm}
\paragraph{Expert agreement.}  
We begin by considering the question \emph{how well do the experts agree w.r.t.~masking potential?} 
This is an important question to consider, as the main task is to emulate the median expert assessment.
Table~\ref{table:f1} shows experts have an \emph{AMAE} ranging from 0.68 to 1.04.
This suggests that, on average, individual experts are almost $\pm1$ masking levels distant from the ground truth -- a reasonable level of agreement.  
A more nuanced picture of expert agreement is given in Figure \ref{fig:network performance}.
Here, agreement between each expert, as well as the ground truth, is measured by {\kendall}.
As a rule-of-thumb {\kendall} $\geq$ 0.3 indicates a strong association \footnote{We refer the reader \href{http://polisci.usca.edu/apls301/Text/Chapter\%2012.\%20Significance\%20and\%20Measures\%20of\%20Association.htm}{here} for an interpretation of {\kendall}.}.
According to this rule, all experts have a strong association, although we can see that experts 1, 2, and 5 exhibit substantially higher agreement than experts 3 and 4.
Interestingly, the experts who tended to agree more were also less experienced.
%, and consequently dominated the ground truth (which is computed as the median).
Turning to the F1-scores in Table~\ref{table:f1}, it is clear that experts are in better agreement for low-masking cases than for high-masking cases.
This suggests that high-masking potential is a generally less \emph{agreeable} property than low-masking potential.
Our findings on the public test set are mirrored in the private test set, provided in Appendix \ref{sec:agreement_private}.
Examples of mammograms where experts agree and disagree are provided in Figure~\ref{fig:discussion}.

\begin{table}[t]
\centering
% \caption{$\mathit{F_1}$ scores on extreme levels on test set}\label{table:f1}
\caption{Comparison of expert and model performance on ordinal classification of masking potential for the public test set.
Mean and standard deviation of 5 runs are reported for the models.}
\label{table:f1}
\scriptsize

\begin{tabular}{cccccc}
\toprule
                    &    &  {Kendall's $\mathit{\tau_b}$} $\uparrow$  & {AMAE} $\downarrow$ & {$\mathit{F_1}$ on level 1-2} $\uparrow$ & {$\mathit{F_1}$ on level 7-8} $\uparrow$ \\ \midrule
\multirow{5}{*}{Experts} & Expert 1  & 0.7232 & \textbf{0.6762} &                               0.7940 & 0.6154           \\
                         & Expert 2  & 0.7279 & 0.7167          &  0.7465 &  \textbf{0.6316}        \\
                         & Expert 3  & 0.5450 & 1.0037          & 0.7363 & 0.5200                    \\
                         & Expert 4  & 0.5554 & 1.0390          & 0.5430 & 0.6242                    \\
                         & Expert 5  & 0.6342 & 1.0321          & 0.6885 & 0.5225                    \\ \midrule
\multirow{2}{*}{Models}  & One-hot   & 0.7126 $\pm$ 0.0083 & 0.8108 $\pm$  0.0145   &  0.7855 $\pm$ 0.0136 & 0.5950 $\pm$ 0.0243                  \\
                         & Multi-hot & \textbf{0.7625} \boldsymbol{$\pm$} \textbf{0.0030} & 0.7086 $\pm$ 0.0142 & \textbf{0.8064}  \boldsymbol{$\pm$} \textbf{0.0188} & 0.5571 $\pm$  0.0320           \\
\bottomrule
\normalsize
\end{tabular}
\vspace{-6mm}
\end{table}

%\ks{Show/discuss examples in Figure 4 where experts disagree. Should we say the same findings hold for the private test set?}

%rule-of-thumb

%The table also shows that two experts have \textit{AMAE} of less than 1, and other three have \textit{AMAE} a little larger than 1. This suggests that on average, individual masking level assessments are at most $\pm1$ level distant from the ground-truth. 

%Notice that in almost all evaluation metrics, Experts 1 and 2 have a comparably better performance compared to the other experts.

%Moreover, by examining the high-masking vs. low-masking F1-scores of the experts, it is clear that all of them achieve a much higher F1 in low-masking mammograms. This suggests that the assessments for high-masking levels could be less \textit{agreeable}.

%Figure \ref{fig:network performance} provides a more in-depth analysis of the performance of the models and experts. We can see that generally, Experts 1, 2, and 5 exhibit higher agreement, and Experts 3 and 4 tend to agree less than others, as reflected in both {\kendall} and \textit{AMAE}. Highest agreement could be seen between Experts 1 and 2.

%\yl{to answer the question- can we trust our ground-truth? show some images with low expert agreement}

\vspace{-3mm}
\paragraph{Task 1: Ordinal classification of masking potential.}
Results for \emph{ResNet34 one-hot} and \emph{ResNet34 multi-hot} (5 runs each) are provided along with the expert agreement in Table~\ref{table:f1} and Figure \ref{fig:network performance}.
We can see that the model designed for ordinal classification, \emph{ResNet34 multi-hot}, outperforms the standard one-hot model according to most metrics: {\kendall}, \textit{AMAE}, and low-masking F1.
The two metrics most sensitive to ordinal relations ({\kendall} and \textit{AMAE}) show a large gap between the two models, suggesting multi-hot encoding is more effective in solving our ordinal classification problem.
The only metric where the one-hot model dominates is the high-masking F1-score, although both seem to struggle (as do the experts).
This comes as something of a disappointment, as high-masking patients are the most interesting from a clinical perspective.
On the other hand, both models performed excellently at identifying low-masking mammograms, and outperformed all the experts.

The models seem to be more correlated with Experts 1, 2, and 5, who agreed with each other more often (and the ground truth). 
Interestingly, our models were more correlated with each individual radiologist
than any of their colleagues were (see the two rightmost columns in Figure \ref{fig:network performance}).
Please note that the cross-tabulation in Figure \ref{fig:amae} is \textit{asymmetric} because the number of mammograms placed into each masking level is different for each radiologist -- \ie the \textit{AMAE} between two experts changes depending on which one is considered as the reference.
In Figure~\ref{fig:discussion}, we provide several examples where the networks both agree and disagree with the experts.
Figure~\ref{fig:discussion}\emph{(g)} is an interesting case because the density is fairly low but experts rate it as high masking potential.
Our models under-estimate the masking, suggesting they rely too heavily on general density cues.

\begin{table}[t]
\caption{AUC on downstream clinical tasks public and private test sets combined.}\label{tab:auc}
\vspace{-3mm}
\begin{center}
\scriptsize
\begin{tabular}{cccc}
\toprule
\multirow{1}{*}{} &  \multicolumn{3}{c}{AUC} \\ %\cline{2-4} 
& \multicolumn{1}{c}{Interval cancer}  & \multicolumn{1}{c}{Large invasive cancer} & \multicolumn{1}{c}{CEP} \\ \midrule
Percent density & 0.5947  & 0.5254 & 0.5678 \\
Dense area & 0.5901  & 0.5505 & 0.5839 \\
One-hot & {0.6321}  $\pm$ {0.0031}  & {0.5801} $\pm$ {0.0013} & {0.6100} $\pm${0.0013} \\
Multi-hot & 0.6331  $\pm$ 0.0031  & 0.5802 $\pm$ 0.0019 & 0.6117 $\pm$ 0.0028 \\
% Multi-hot & $\SI{0.6306}{} \pm \SI{0.0021}{}$  & $\SI{0.5865}{} \pm \SI{0.0029}{}$ & $\SI{0.6153}{} \pm \SI{0.0031}{}$ \\
\bottomrule
\normalsize
\end{tabular}
\end{center}
\vspace{-5mm}
\end{table}

\begin{figure}[t]
\hspace{-1mm}
\begin{tabular}{@{}c@{\hspace{3mm}}c@{}}
{\includegraphics[width=0.49\columnwidth]{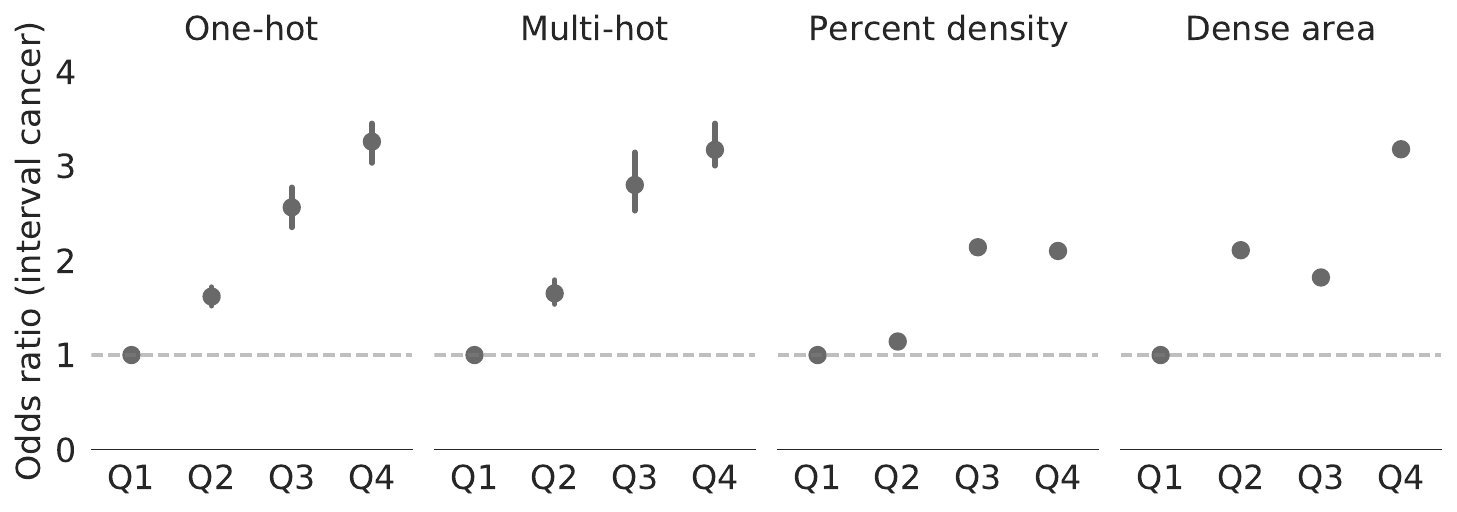}} &
{\includegraphics[width=0.49\columnwidth]{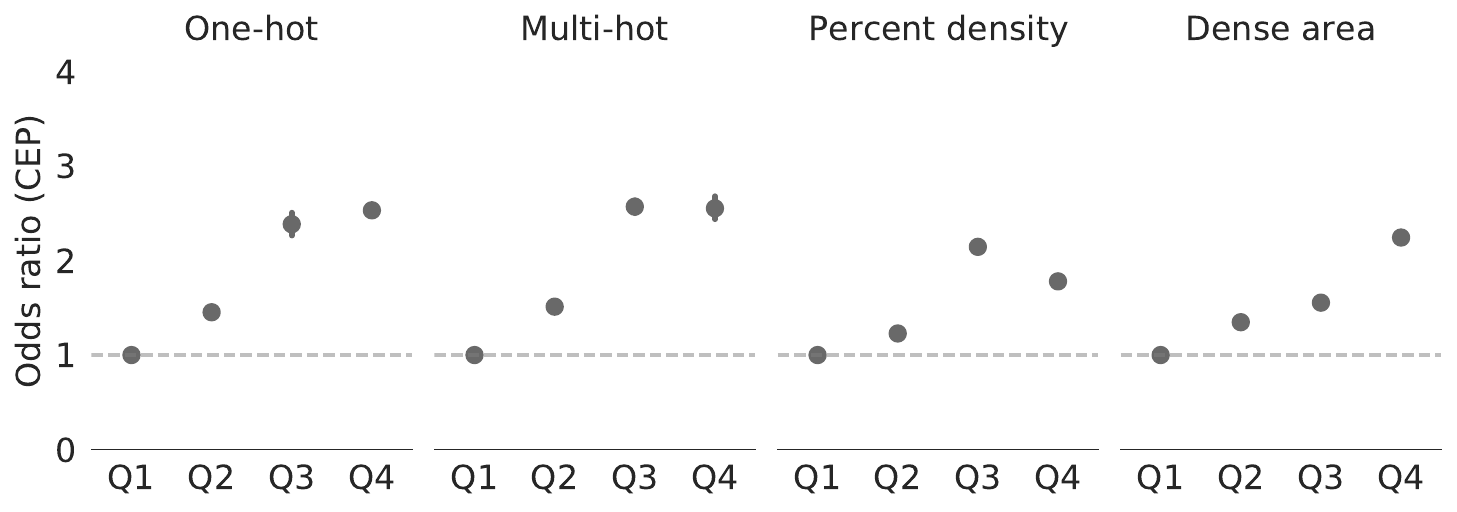}} \\
\emph{(a)} Odds ratio (interval cancer)& 
(b) Odds ratio (CEP) \\
\end{tabular}
\vspace{-2mm}
\caption{Odds ratio on clinical endpoints with public and private test sets combined.} 
\label{fig:odds ratio plot}
\vspace{-4mm}
\end{figure}

% \begin{figure}[t]
% \centering
% \subfloat[Odds ratio (interval cancer)\label{fig:or_interval}]{\includegraphics[width=0.48\columnwidth]{figs/experiments/oddsratio_interval.pdf}}\quad
% \subfloat[Odds ratio (CEP)\label{fig:or_composite}]{\includegraphics[width=0.48\columnwidth]{figs/experiments/oddsratio_composite.pdf}}\quad
% \caption{Odds ratio on clinical endpoints with public and private test sets combined.} 
% \label{fig:odds ratio plot}
% \vspace{-5mm}
% \end{figure}

\vspace{-3mm}
\paragraph{Task 2: Identification of interval and large invasive cancers.}
We set out to investigate if estimates of masking potential are predictive of \emph{(1)} interval cancer, \emph{(2)} large invasive cancer, and \emph{(3)} the composite endpoint (CEP), \ie interval \emph{or} large invasive cancer.
There is reason to believe a correlation exists, because interval cancers or cancers that appear to grow fast can result from a misdiagnosed mammogram, which is more likely if masking potential is high.
Due to the low number of cancers randomly sampled into the public test set, this part of the analysis is performed on the combined public and private test sets for increased statistical power.
%\yl{To get more statistical power, this part of analysis was done with combined public and private test sets.}
In Table \ref{tab:auc} we compare our models against \emph{percent density} and \emph{dense area} at identifying these outcomes, as measured by AUC.
Both models are stronger indicators than the density measures by a large margin, with \emph{ResNet34 multi-hot} slightly outperforming \emph{ResNet34 one-hot}.
%\ks{add statistical tests to Table 4.}
All methods perform stronger at predicting interval cancer than large invasive, which makes sense because the relation to masking is more likely to be causal -- \eg a missed diagnosis caused by masking.

In Figure \ref{fig:odds ratio plot} we plot the \emph{odds ratio} for successive quartiles of the predictions from various models.
Recall that good clinical predictors should exhibit monotonically increasing odds ratios, with strong odds ratios in the highest quartiles. 
If masking is related to interval cancer, then a high-masking prediction should have much higher odds to identify a cancer than a low-masking prediction.
We can see that this is the case for both \emph{ResNet34} models, with odds of finding a interval cancer 3.2 times higher in the top quartile than the first.
In contrast, the density measures yield lower odds ratios and fail the monotonicity test, indicating they do a poor job of sorting cancer risk.
The odds ratios for large invasive cancers, given in the Appendix \ref{sec:odds_ratio_invasive}, are less promising.
This trend is also evident in the AUC results from Table \ref{tab:auc}, suggesting that identifying screening participants confirmed with large invasive cancer is more challenging than interval cancer.
The composite endpoint, which combines both cancers, reflects this in the AUC and odds ratios.

\section{Conclusions}\label{sec:conclusions}

\vspace{-3mm}
There is a strong clinical interest in predicting interval and large invasive cancers, as they indicate a failure of the screening infrastructure and may lead to poor prognoses.
%The former indicate a failure of the existing screening infrastructure, and the latter is likely to be lethal.
We have shown that deep learning models trained on CSAW-M can identify these cancers significantly better than density measures.
This suggests that the expert masking potential information provided by CSAW-M has high clinical value, aside from its value as an ordinal classification benchmark and the other merits listed in the introduction.
Our baseline models succeeded at modeling masking potential, agreeing with the ground truth better than any individual expert.
However, agreement among experts and our models was much better for low-masking than for high-masking.
We note that our baselines are very simple models that were not explicitly trained to learn the clinical outcomes.
As such, there is much potential for improvement.

%List of some concluding points:
%\begin{itemize}
%    \item Annotator agreement much better for low-masking than high-masking
%    \item CSAW-M is multi-purpose, it is annotated with masking level, but a model trained to assess masking could be used \textit{off-the-shelf} in other tasks e.g. clinical outcomes
%\end{itemize}

%\vspace{-3mm}
%\paragraph{Limitations}

%\ks{discuss limitations in Conclusion and broader impact, adjust datasheet accordingly}

\vspace{-3mm}
\paragraph{Limitations and broader impact.}
%\section*{Conclusion and broader impact}
CSAW-M is primarily intended for the development of automated systems to estimate masking of mammograms. 
Such a system can be used in the screening process to pre-emptively detect interval and large invasive cancers, as we are attempting to show in an ongoing clinical trial, ScreenTrust MRI\footnote{\url{https://www.clinicaltrials.gov/ct2/show/NCT04832594}}. 
Furthermore, since CSAW-M is the largest public mammography dataset, it can enable more fruitful research and development for different applications that involve automatic analysis of mammograms, for instance, through unsupervised learning. 
Such automated systems can help with the world-wide shortage of specialists. On the other hand, a general concern when releasing a human cohort dataset is malicious use of the released data to re-identify the individuals. 
Therefore, we have taken measures to mitigate this issue: (1) we removed all individual identifiers from the data, (2) we down-sampled the mammograms, (3) we removed all unnecessary acquisition attributes --DICOM headers--, (4) we simplified the continuous tumor size attribute to a binary outcome, and (5) we anticipated a gated release mechanism to approve users based on their information and project goals before granting access to the data. 
It should be noted, however, that while these efforts make re-identification extremely unlikely, it does not provide a theoretical guarantee. 
We acknowledge several limitations  and biases present in CSAW-M. 
For example, there are no visible tumors in CSAW-M by design.
This limits its usefulness for tumor detection.
Also, the training set is limited to a single annotation per image, as a compromise between annotation cost and number of examples.
This affects the training data more than the test data, as the ground truth is the median of 5 expert opinions.
Biases are introduced from a number of sources, \eg the training and background of the experts.
We should also be aware of biases inherent in the collection process, as it was collected from a certain population, period, and region, using certain imaging equipment (\eg we selected more screening participants with cancers than are present in the general population).
To alleviate this, the \textit{known} specifics of the population and our selection pipeline are thoroughly described in the paper.
Nevertheless, clinical studies are crucially required before deploying models in any clinical processes.

\vspace{-3mm}
\paragraph{Acknowledgements.}
This work was supported by the Regional Cancer Center Stockholm-Gotland as source for the clinical data and partly financed by MedTechLabs \url{https://www.medtechlabs.se/}, the Swedish Research Council (VR) 2017-04609, and Region Stockholm
HMT 20200958.
We thank Christos Matsoukas and Johan Fredin Haslum for an early review of this work.

%\subfile{sections/5_discussions}
%\subfile{sections/6_conclusions}
%\subfile{sections/7_final_remarks}

%\begin{ack}
%Use unnumbered first level headings for the acknowledgments. All acknowledgments
%go at the end of the paper before the list of references. Moreover, you are required to declare
%funding (financial activities supporting the submitted work) and competing interests (related financial activities outside the submitted work).
%More information about this disclosure can be found at: %\url{https://neurips.cc/Conferences/2021/PaperInformation/FundingDisclosure}.

%Do {\bf not} include this section in the anonymized submission, only in the final paper. You can use the \texttt{ack} environment %provided in the style file to automatically hide this section in the anonymized submission.
%\end{ack}

% \bibliographystyle{plainnat}
\bibliographystyle{splncs04}
\bibliography{reference}

\newpage
\appendix
\section*{Appendix}

\section{Image selection procedure flowchart}
\label{sec:image_selection_procedure}
A flowchart detailing the data selection process for the creation of CSAW-M.

\begin{figure}[h!]
\centering
\includegraphics[width=0.7\textwidth]{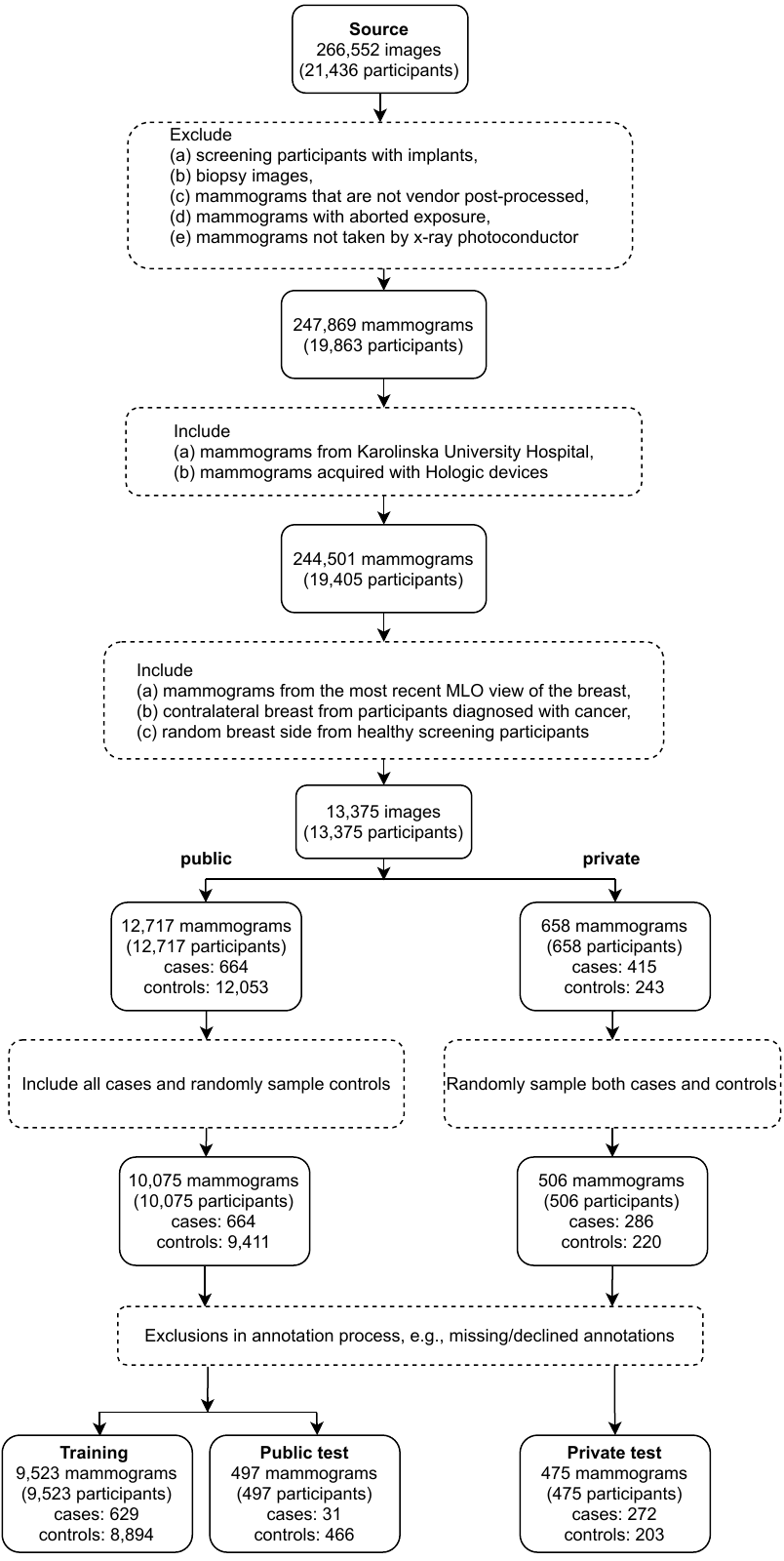}
% \vspace{-3mm}
\caption{Data selection flowchart.}
\label{appendix:dataset_flowchart}
%\vspace{-5mm}
\end{figure} 

\section{Sampling screening participants and pre-processing of images in CSAW-M}
\label{sec:sampling_and_preprocessing}
\begin{figure}[h]
    \centering
    \subfloat[The private test set]{\includegraphics[scale=.4]{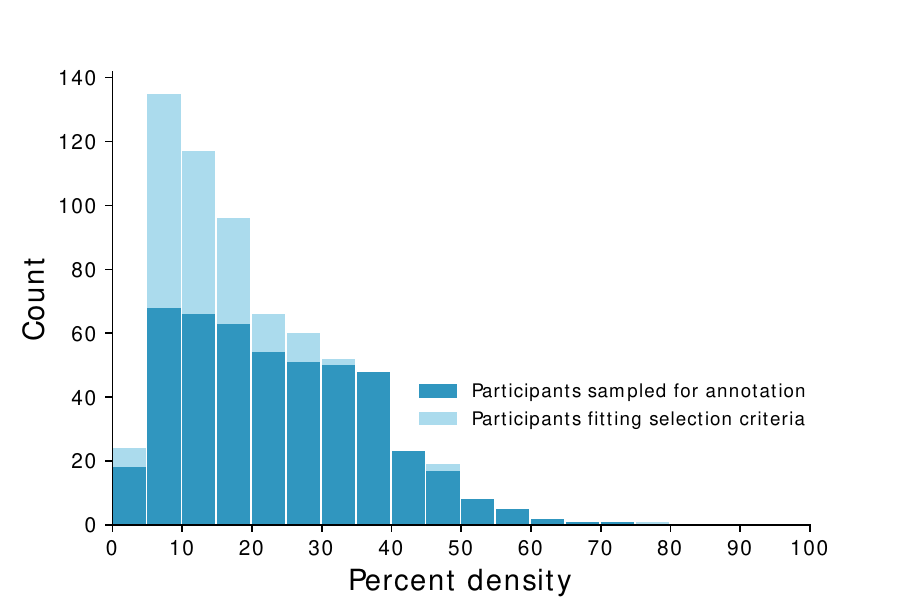}\label{fig:private_sampling_dist}}
    \subfloat[The public training and test sets]{\includegraphics[scale=.4]{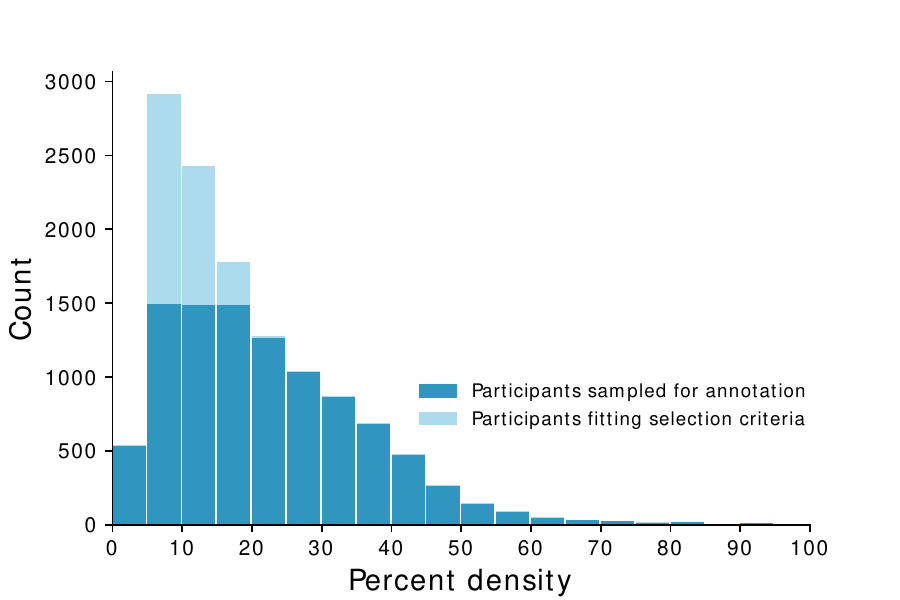}\label{fig:public_sampling_dist}}
    \caption{Distribution of breast percent density for different splits of CSAW-M.}
    \label{fig:sampling_dists}
\end{figure}

\paragraph{{Sampling screening participants for CSAW-M by percent density}.}
In order to have images with diverse breast densities, we uniformly sampled images with intermediate densities while keeping all images in the tails. The percent densities are calculated with the publicly available software Libra \cite{keller2015preliminary}. 
Figure \ref{fig:sampling_dists} shows the resulting distribution of breast density in the public and private portions of CSAW-M. 
%Note that even though we kept all images in the tails, the number of women in CSAW-M in that some tail density intervals may be lower than women fitting selection criteria (e.g. women with breast density below 5 Figure \ref{fig:private_sampling_dist}) \yl{Moein, this sentence is unclear}. This is because some of the images were discarded by the experts during the annotations (due to e.g. low quality of the image, an incomplete view of the breast etc.)

\paragraph{Finding and removing the area in the image that contains text.} We remove the text in the image using the OpenCV library \cite{opencv_library}. We try to find the contour which includes text in the image, and then replace the pixels within that contour. Using a binarized copy of the image (in which each pixel greater than 0 is mapped to 255 and pixels with value 0 remain 0, resulting in a totally black and white copy of image), we first extract all available contours. Note that the letters themselves are among the contours.
We ignore the largest contour in the image - which corresponds to the breast and consider other available contours as candidates. This ensures that the breast remains intact. Since we know that the letters in the image are spatially close to each other, we apply a dilation transformation to the image so we could combine the little adjacent contours in the image. This transformation essentially combines smaller contours corresponding to letters into a larger contour. We also ignore all the contours below the breast, since we know beforehand that the text will never be there. 
Among the remaining contours, we finally only consider the ones whose area is at least greater than a threshold (which we found by trial and error). If there are multiple contours remaining, we choose the one that is closest to the top-right corner of the image, since we already know the text is almost always around there. This will result in the contour containing the text in the image.
Once the contour including the text is found, we use its coordinates and replace the values inside it in the original image with the minimum pixel value that is available in the original image (which is usually black, corresponding to air).

\section{Details on the annotation procedure}
\label{annotation_procedure_details}

\begin{figure}[h]
    \centering
    \includegraphics[scale=0.3]{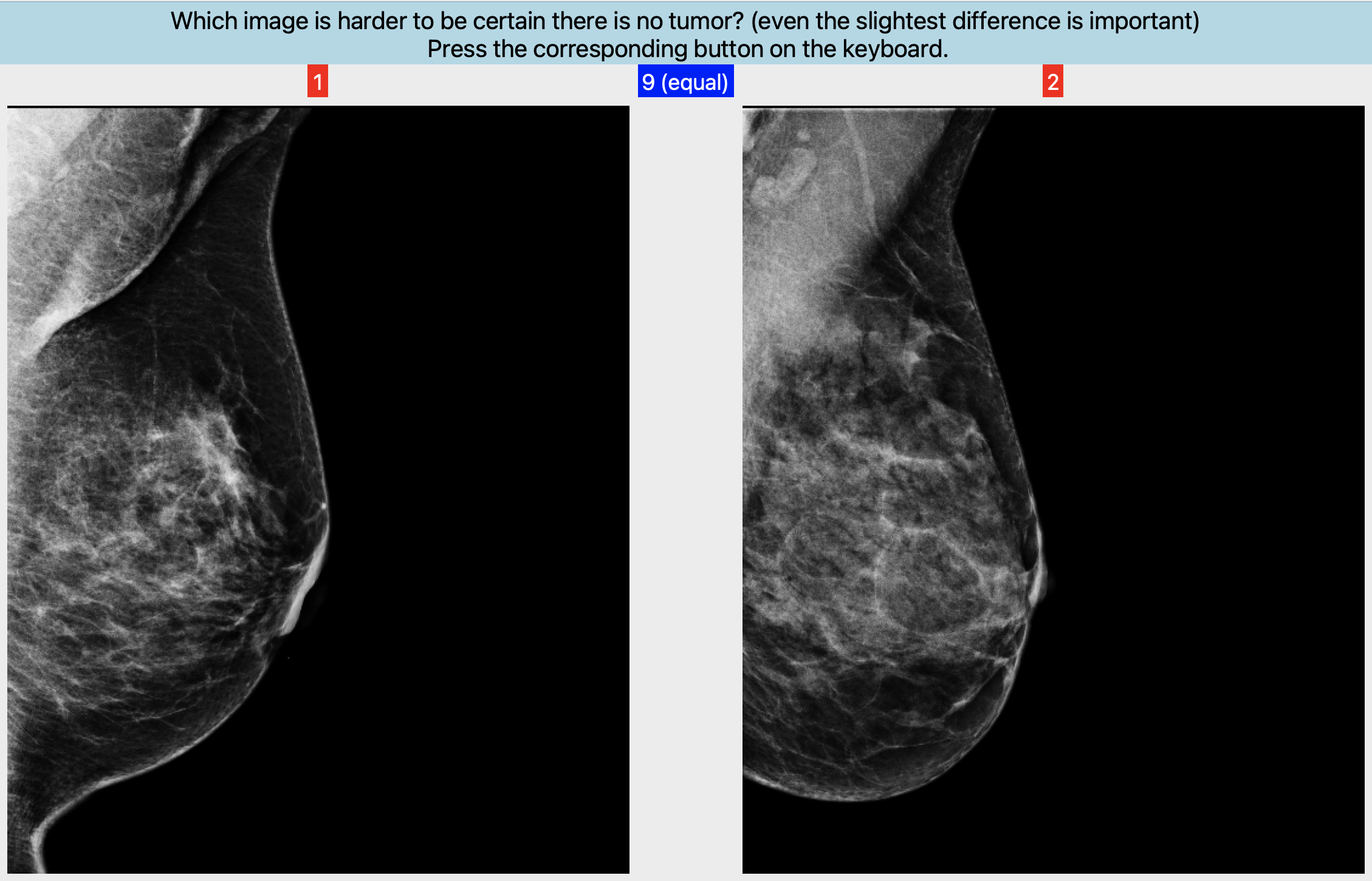}
    \caption{Interface of our annotation tool. 
    See text for details.}
    %In order to make the comparison easier for the annotators, images are flipped if necessary so that both breasts point to the same direction, making the assessment easier. We always showed the \textit{query image} on the left and the \textit{reference image} on the right. We asked the annotators to press '1' is they evaluate that the left image is harder to assess (exhibiting a higher masking level), '2' if the right image is harder to assess, and '9' if they see no discernible difference between them. At any point in time, annotators were provided the option to \textit{discard} the current pairwise in case they could not make a reasonable judgement, \eg because the query image is distorted, it does not contain the whole breast etc.)}
    %\ms{This would be moved to appendix, explaining that all the images in the project are left-posed but only in annotation interface they were shown right-posed. The bottom section of the the interface is redundant (case number and below...) The question in the interface should be changed.}}
    \label{fig:annotation_interface}
\end{figure}

\paragraph{Annotation tool.} 
We developed an annotation tool and distributed it to the experts for annotating CSAW-M.
The tool displayed two mammogram for a \emph{pairwise comparison} and asked the expert: \emph{which image is harder to be certain there is no tumor?}.
The interface is shown in Figure~\ref{fig:annotation_interface}.
In order to make the comparison easier for the annotators, images are flipped if necessary so that both breasts point to the same direction, making the assessment easier. We always showed the \textit{query image} on the left and the \textit{reference image} on the right. We asked the annotators to press ``1'' is they evaluate that the left image is harder to assess (exhibiting a higher masking level), ``2'' if the right image is harder to assess, and '9' if they see no discernible difference between them. At any point in time, annotators were provided the option to \textit{discard} the current pairwise in case they could not make a reasonable judgement, \eg because the query image is distorted, it does not contain the whole breast etc. The tool was used for both binary and ternary search without any changes to the user experience. 

\begin{figure}
    \centering
    \includegraphics[scale=0.5]{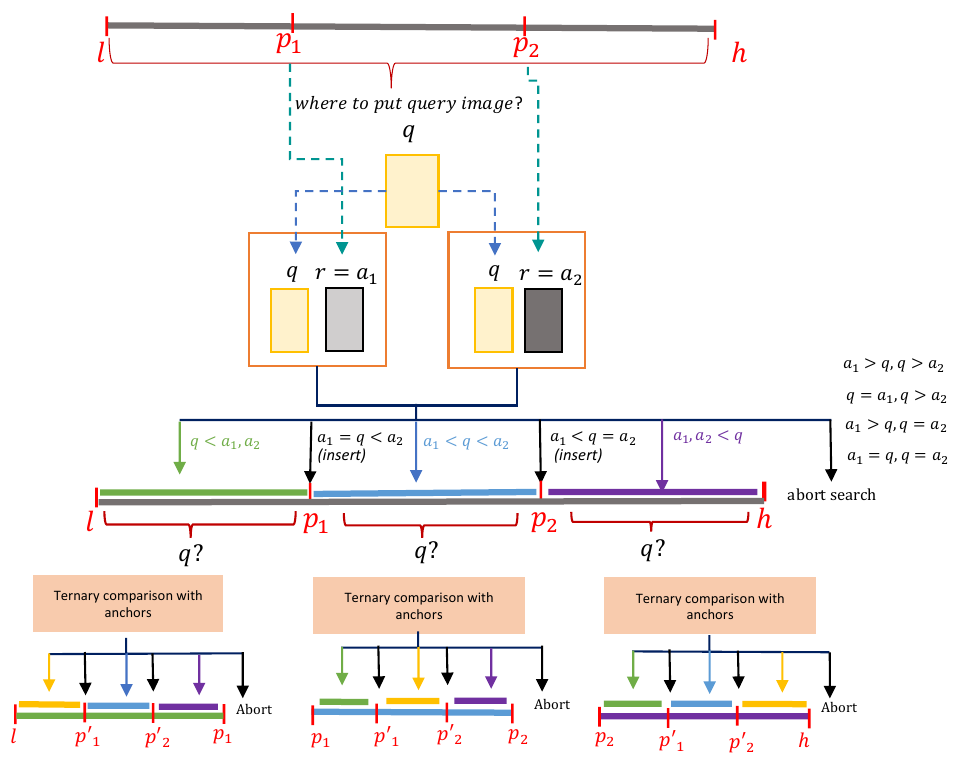}
    \caption{Illustration of the pairwise comparisons used in our ternary search, where the query image is compared against two anchors from the sorted list. Based on the assessment from the annotator, either the ternary search interval gets updated, insertion happens, or the search is aborted.}
    \label{fig:ternary}
\end{figure}

\paragraph{Technical details about the ternary comparisons for the private test set}
There are 9 possibilities for how the query image $q$ could be assessed against anchors $a_1$ and $a_2$. Figure \ref{fig:ternary} illustrates different possibilities and show the outcome of each of them. We consider the two consecutive comparisons \textit{consistent} if: 1) $q > a_1, a_2$ 2) $q < a_1, a_2$ or 3) $a_1 < q < a_2$.
In the case of consistent comparisons, the search interval will be updated accordingly, and the search will continue to its next level. We directly insert $q$ at position $p_1$ if $a_1 = q < a_2$ and we insert at position $p_2$ if $a_1 < q = a_2$. All other possibilities are considered to be \textit{inconsistent}, in which case the search is \textit{aborted}, and we start the search from the beginning for a new query image. At the end of the annotation procedure, we asked the experts to make new attempts to insert the aborted images into the list. Since the sorted list would continuously get updated after inserting new images, the new attempts would usually result in successfully inserting the images whose search was previously aborted. Note that in the new attempts, the search process for inserting the query images would begin from scratch.

\paragraph{Technical details about the binary search for the private test set} Given the search interval $\left[l, h\right]$, the image lying at position $p=\left(l+h\right)/2$ would be selected as the \textit{reference} image. The search interval is updated based on the result of the pairwise comparison. If $q > r$ the search interval would change to $\left[p+1, h\right]$, if $q < r$ the search interval would change to $\left[l, p-1\right]$, and if $q = r$ the image would get directly inserted at position $p$.

\section{Technical details on the models}
\label{details_on_models}
\paragraph{Implementation details} We used the PyTorch framework \cite{NEURIPS2019_9015} for training our networks. We also used the scikit-learn \cite{scikit-learn} and SciPy \cite{2020SciPy-NMeth} libraries for computing the metrics we used in our experiments. Training each model was done on a single NVIDIA Quadro RTX 8000, taking $\sim$2 hours to complete.
\paragraph{Calculating the continuous masking score} 
We can aggregate the output probabilities of a model into a single continuous score that we call \textit{masking score}. For the \textit{one-hot} model, we simply compute the score as a weighted average of probabilities as: 
\begin{align}
    \label{score_formula}
    s = \frac{1}{8}(\sum_{l=1}^8 \mathit{p}_l \cdot l)
\end{align}

where $p_l$ represents the probability that the input image belongs to masking level $l$.

%where $P$ represents the softmax output probabilities $p_l \in P$ and $y_l$ corresponds to the label for masking level $l$, being either 0 or 1 following one-hot encoding scheme. 

To get continuous scores from the \textit{multi-hot} model, we first need to convert the cumulative probabilities that each output head produces to actual masking level probabilities $p_l$. Recall that the multi-hot model trained for ordinal classification with 8 classes would have 7 output heads. As mentioned in the main text, each output head models an individual binary classifier. For an input image $x$ with true label $L$, each output head \textit{independently} produces $P_k = p\left(L > k\right)$ where $k \in \{1, 2, .., 7\}$, denoting the (cumulative) probability that $L$ exceeds $k$. Since in general each output head is an independent binary classifier, the output probabilities are not guaranteed to be \textit{monotonic} (although most often they are), so we apply a small trick to make them monotonic, \textit{i.e.,} for each $k' > k$, it holds that $P_{k'} <= P_k$. This is implemented as in the code snippet below.

% \begin{minted}{python}
\begin{lstlisting}
    def make_monotonic(cdf_list):
        monotonic = []
        for i in range(7):
            max_cdf = max(cdf_list[i:])
            monotonic.append(max_cdf)
        return monotonic
\end{lstlisting}

Once the probabilities are made monotonic, individual masking level probabilities could be calculated by subtracting consecutive cumulative probabilities such that for each $k' = k + 1$ it holds that $p_k = P_k - P_{k'}$, as shown in the following code snippet:
% \begin{minted}{python}
\begin{lstlisting}
    def make_probs(lst):
        extended = deepcopy(lst)
        extended.insert(0, 1)  # cdf: 1 in the beginning
        extended.append(0)  # cdf: 0 at last
        probs = []
        for i in range(0, len(extended) - 1):
            probs.append(extended[i] - extended[i + 1])
        return probs
\end{lstlisting}
% \end{minted}

Once we have evaluated individual masking probabilities, we could use the same formula as in Equation \ref{score_formula} to calculate the continuous score.

\section{Distribution of ground-truth and annotations from each expert }\label{sec:experts_dist}

\begin{figure}[h]
\centering
\subfloat[GT]{\includegraphics[width=0.25\columnwidth]{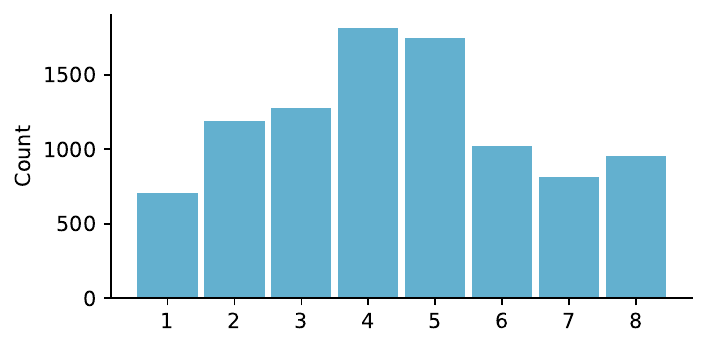}}\quad
\subfloat[Expert 1]{\includegraphics[width=0.25\columnwidth]{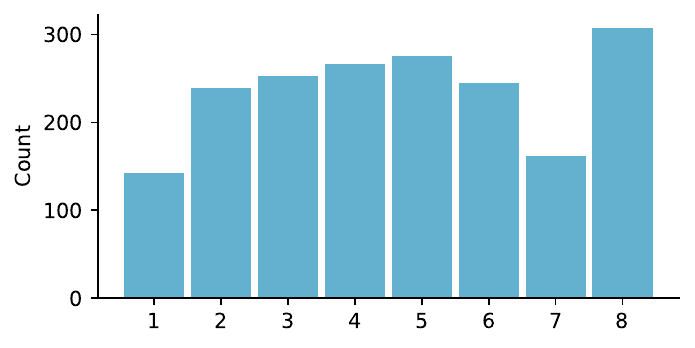}}\quad
\subfloat[Expert 2]{\includegraphics[width=0.25\columnwidth]{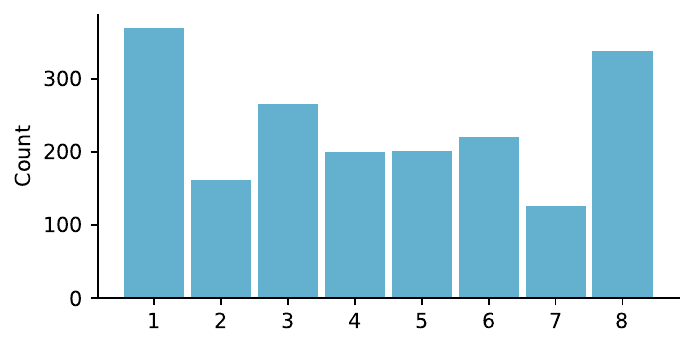}}\quad \\

\subfloat[Expert 3]{\includegraphics[width=0.25\columnwidth]{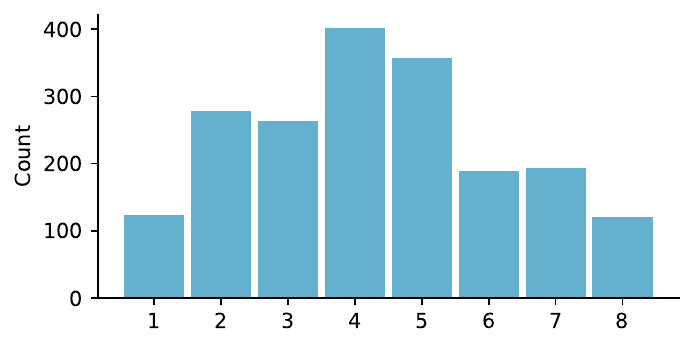}}\quad
\subfloat[Expert 4]{\includegraphics[width=0.25\columnwidth]{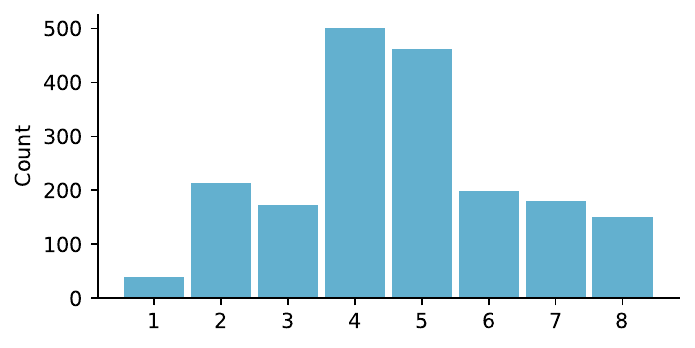}}\quad
\subfloat[Expert 5]{\includegraphics[width=0.25\columnwidth]{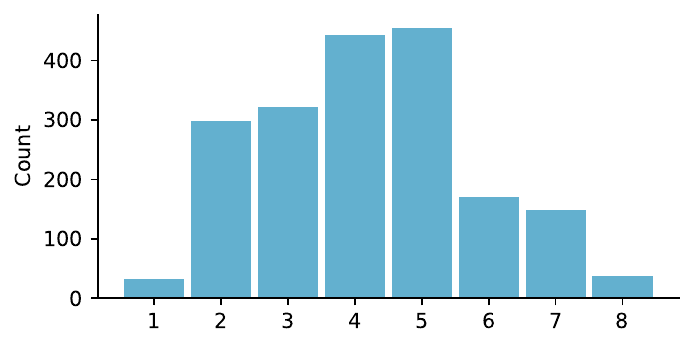}}\quad \\
\caption{Distribution of ground-truth and annotations from each expert on public training set. } 
\label{fig:annotation_hist_train}
\end{figure}

\begin{figure}[h]
\centering
\subfloat[GT - Median]{\includegraphics[width=0.25\columnwidth]{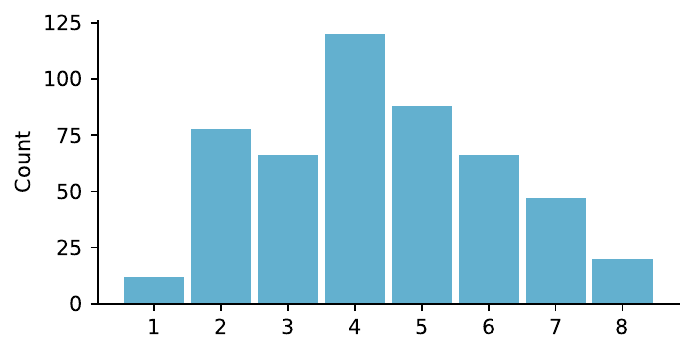}}\quad
\subfloat[Expert 1]{\includegraphics[width=0.25\columnwidth]{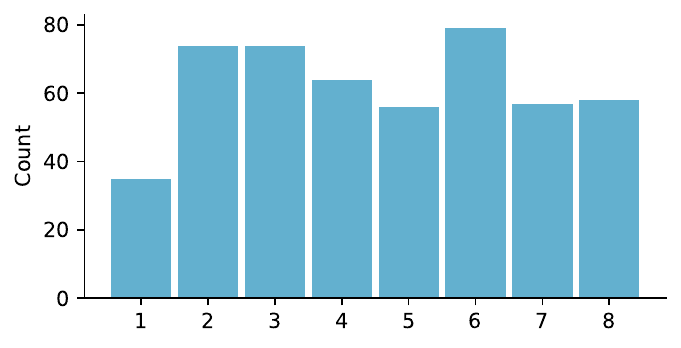}}\quad
\subfloat[Expert 2]{\includegraphics[width=0.25\columnwidth]{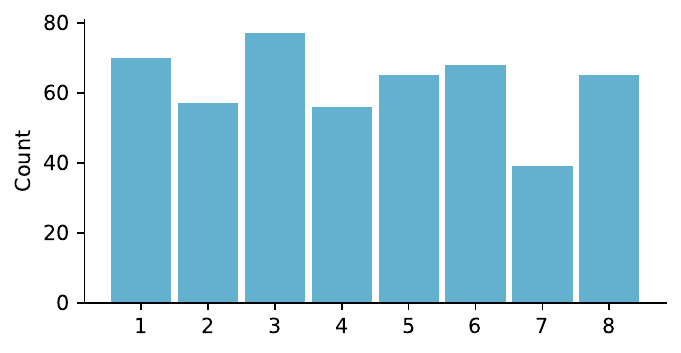}}\quad \\

\subfloat[Expert 3]{\includegraphics[width=0.25\columnwidth]{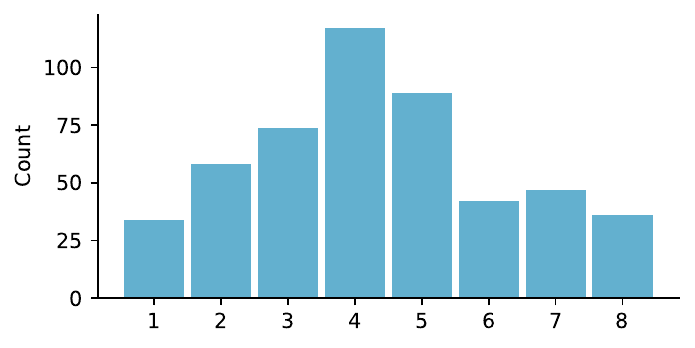}}\quad
\subfloat[Expert 4]{\includegraphics[width=0.25\columnwidth]{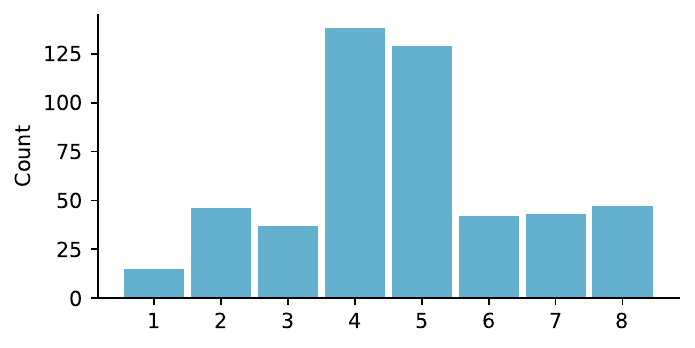}}\quad
\subfloat[Expert 5]{\includegraphics[width=0.25\columnwidth]{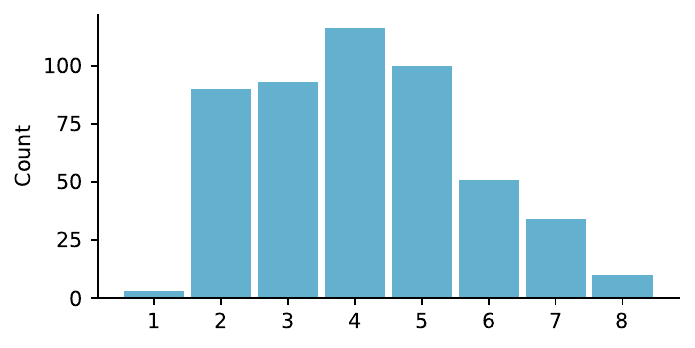}}\quad \\
\caption{Distribution of ground-truth and annotations from each expert on public test set. } 
\label{fig:annotation_hist}
\end{figure}

In Figure~\ref{fig:annotation_hist_train} and Figure~\ref{fig:annotation_hist} we provide the distributions of masking level annotations for each expert on the public training and test sets. 
The distribution of ground truth annotations is also provided. 
The ground truth for each test image is the median of expert annotations.

\section{Ordinal classification of masking potential on private test set}\label{sec:agreement_private}
\begin{figure}[h!]
\centering
\subfloat[Kendall's $\mathit{\tau_b}$ (higher is better) \label{fig:kendall_private}]{\includegraphics[width=0.4\columnwidth]{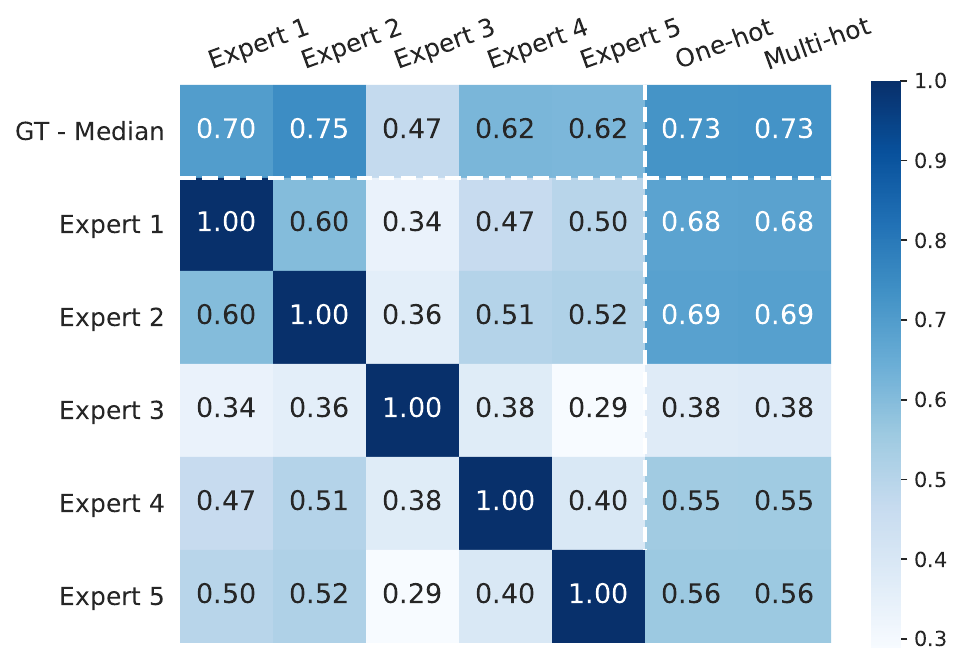}}\quad
\subfloat[{AMAE} (lower is better) \label{fig:amae_private}]{\includegraphics[width=0.4\columnwidth]{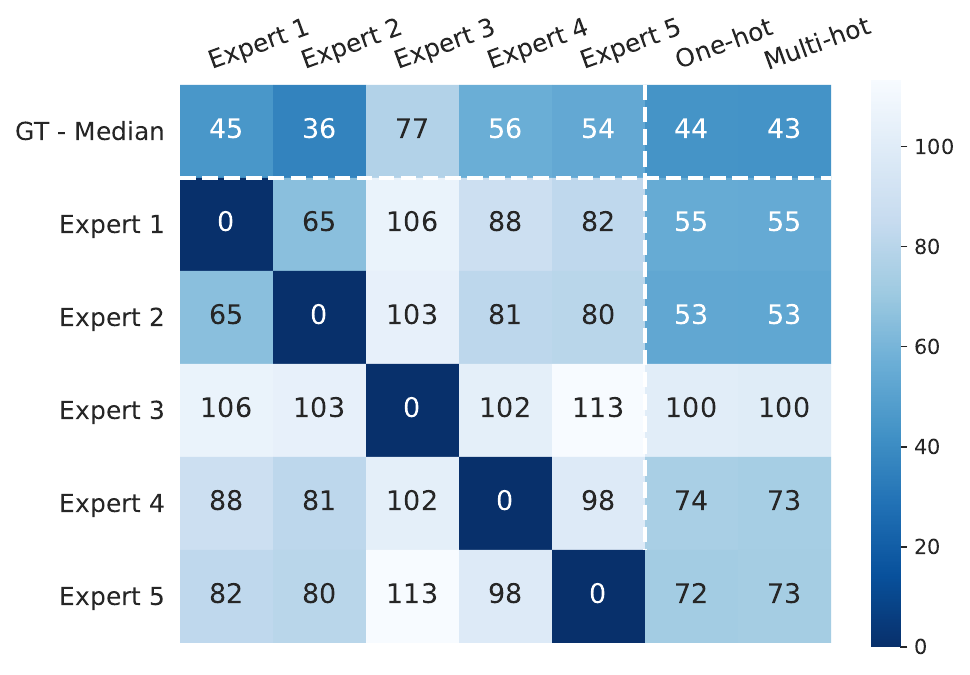}}\quad
\caption{Expert and model agreement on private test set. We perform five runs on our one-hot and multi-hot models and report the mean. } 
\label{fig:network performance private}
\end{figure}

Model and expert agreement on the private test set is shown in Figure \ref{fig:network performance private} with {\kendall} and \textit{AMAE}. Similar conclusions can be drawn in private test set as that of public test set in Figure \ref{fig:network performance}.

\section{Identification of large invasive cancer}\label{sec:odds_ratio_invasive}
\begin{figure}[h!]
\centering
\hspace{-1mm}
\begin{tabular}{c}
{\includegraphics[width=0.7\columnwidth]{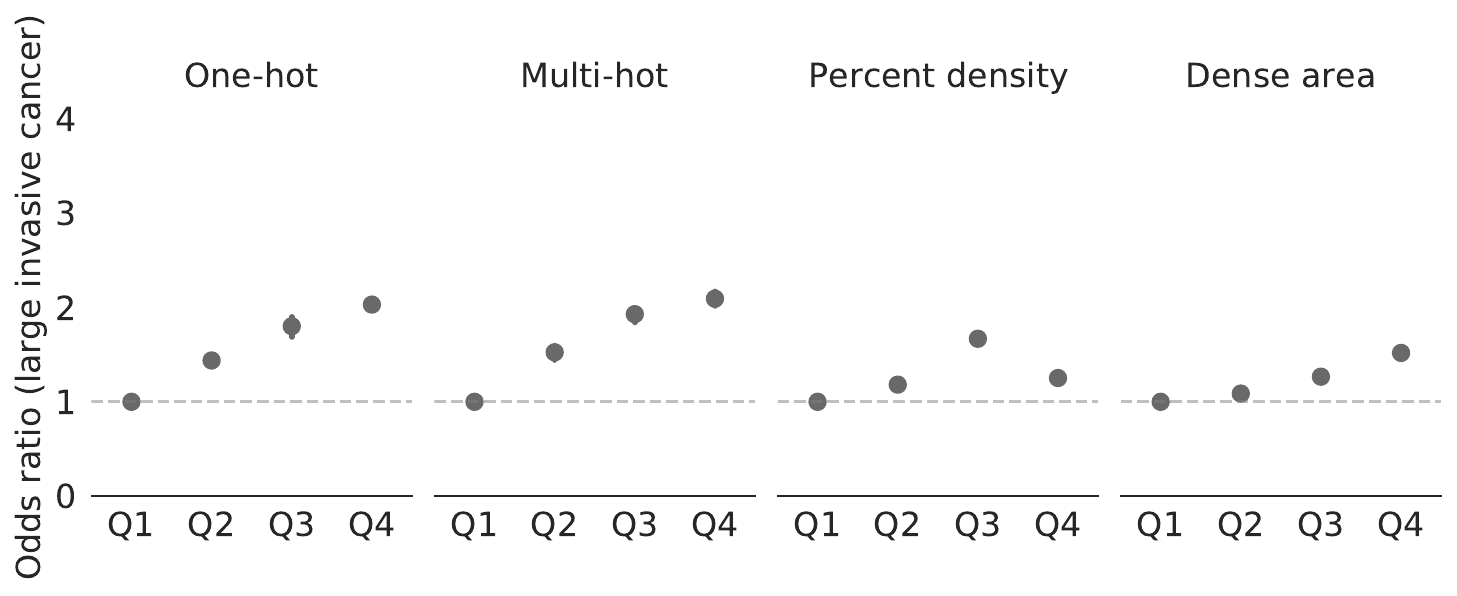}} \\
\end{tabular}
\vspace{-2mm}
\caption{Odds ratio on large invasive cancer with public and private test sets combined.} 
\label{fig:odds ratio plot invasive}
\end{figure}

Figure \ref{fig:odds ratio plot invasive} shows the odds ratio on large invasive cancer with public and private test sets combined. The results are less promising compared to the odds ratio on interval cancer and CEP, shown in Figure \ref{fig:odds ratio plot}.
 
\section{Variations in masking levels}\label{sec:percent_density_variation}

\begin{figure}[h!]
\centering
\begin{tabular}{c}
{\includegraphics[width=\columnwidth]{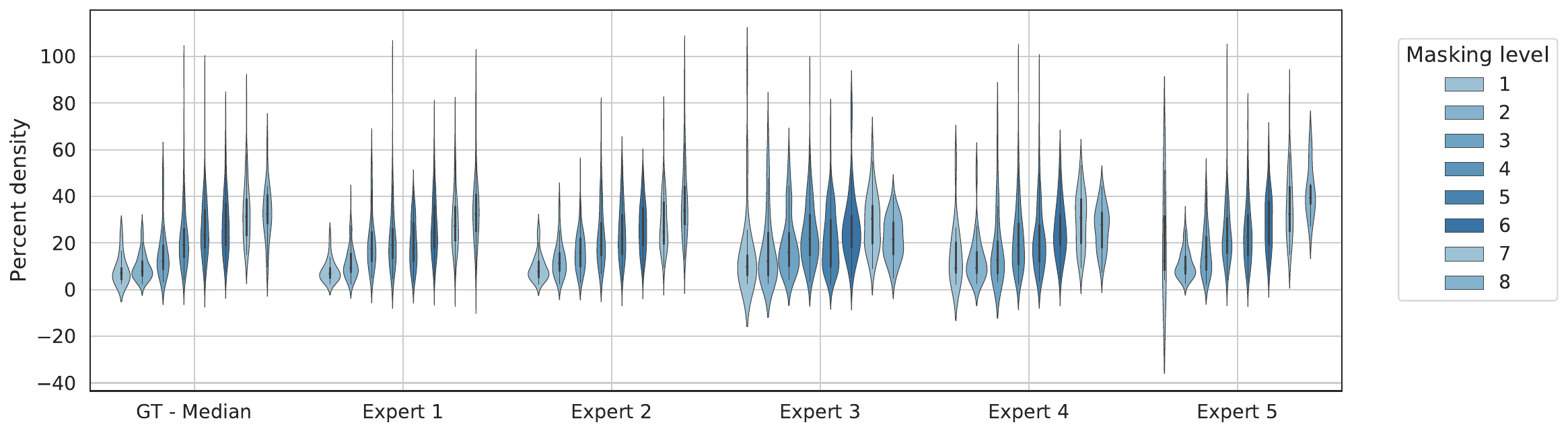}} \\
\end{tabular}
\vspace{-2mm}
\caption{Variations in different masking levels using percent density as a proxy on public test set.} 
\label{fig:violin_pd}
\end{figure}

In Figure \ref{fig:violin_pd}, violin plots show the distribution of percent density as a proxy of variation within each masking level, grouped by expert. We can see that experts 1 and 2, whose masking-level distribution was almost uniform (as seen in Figure~\ref{fig:annotation_hist}), show high variability for high masking levels. This can be explained by the heavy tail of the percent density distribution even after undersampling (as described in Figure \ref{fig:sampling_dists}). Uniform binning would result in more images with diverse percent density (from the tail of the distribution) to be included in their high masking levels. Moreover, the median of the violin plots changes more significantly for these experts as in higher masking levels.

% Similarly, the change in the median of the violin plots for these experts increases for high masking levels.

%Figure \ref{fig:violin_pd} shows that for experts who annotated the images uniformly (namely experts 1 and 2, see Figure \ref{fig:annotation_hist}), we can see a clear increasing trend in the variation of percent density in higher masking levels. We believe this is due to the long tail of percent density distribution even after undersampling (as seen in Figure \ref{fig:before_sampling_hist} of the paper). Experts who made uniform binning would necessarily have images with more diverse percent density values in their levels 7-8, and we can see this is the case considering the spread of their violin plots in levels 7-8 for these experts.
%We can also see that the variation in masking levels 1 and 2 are much lower for both of the experts. With the median, we can see a similar trend, but not as much as experts 1 and 2.

%Although it is hard to exactly quantify the gap between different masking levels, if we consider the median points in violin plots, we can infer that with uniform binning (experts 1 and 2). As shown in Figure \ref{fig:violin_pd}, the difference between medians in masking levels 7-8 is larger than that of their masking levels 3-4, suggesting that the gap between adjacent higher masking levels is higher than that of earlier levels.

\section{Agreements in masking levels}\label{sec:amae_agreement_each_masking_level}

\begin{figure}[h!]
\centering
\subfloat[{AMAE}, median as reference \label{fig:amae_4_maskinggrounps_gt}]{\includegraphics[width=0.45\columnwidth]{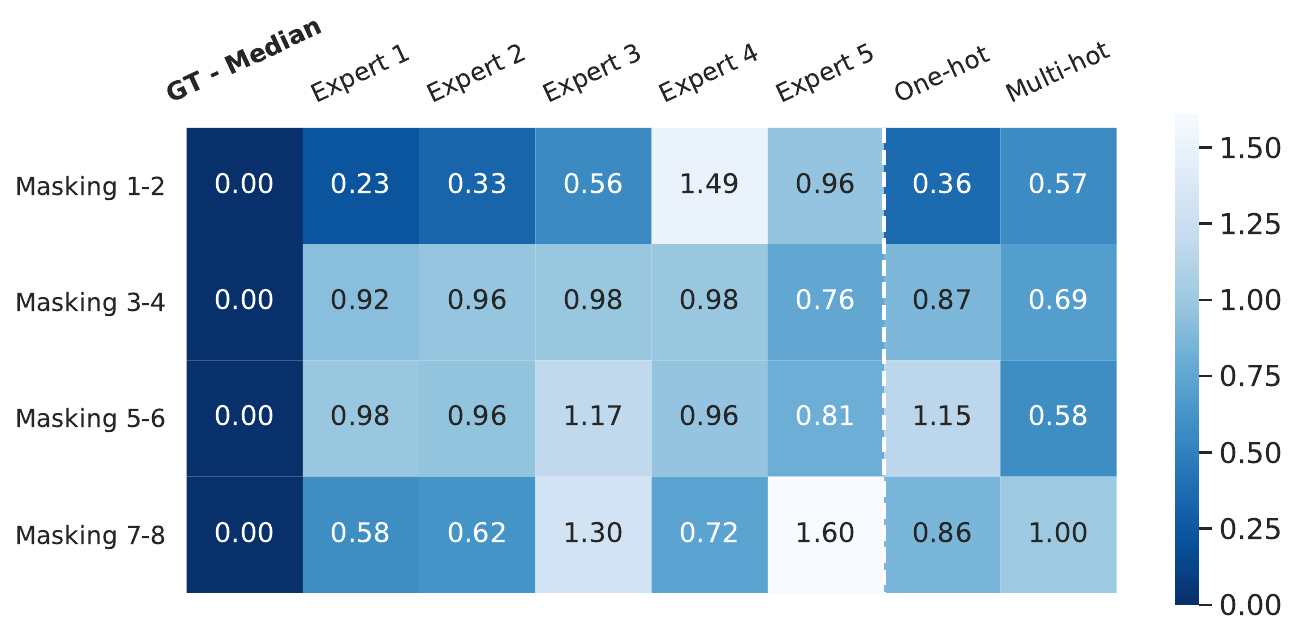}}\quad
\subfloat[{AMAE}, expert 1 as reference
\label{fig:amae_4_maskinggrounps_expert1}]{\includegraphics[width=0.45\columnwidth]{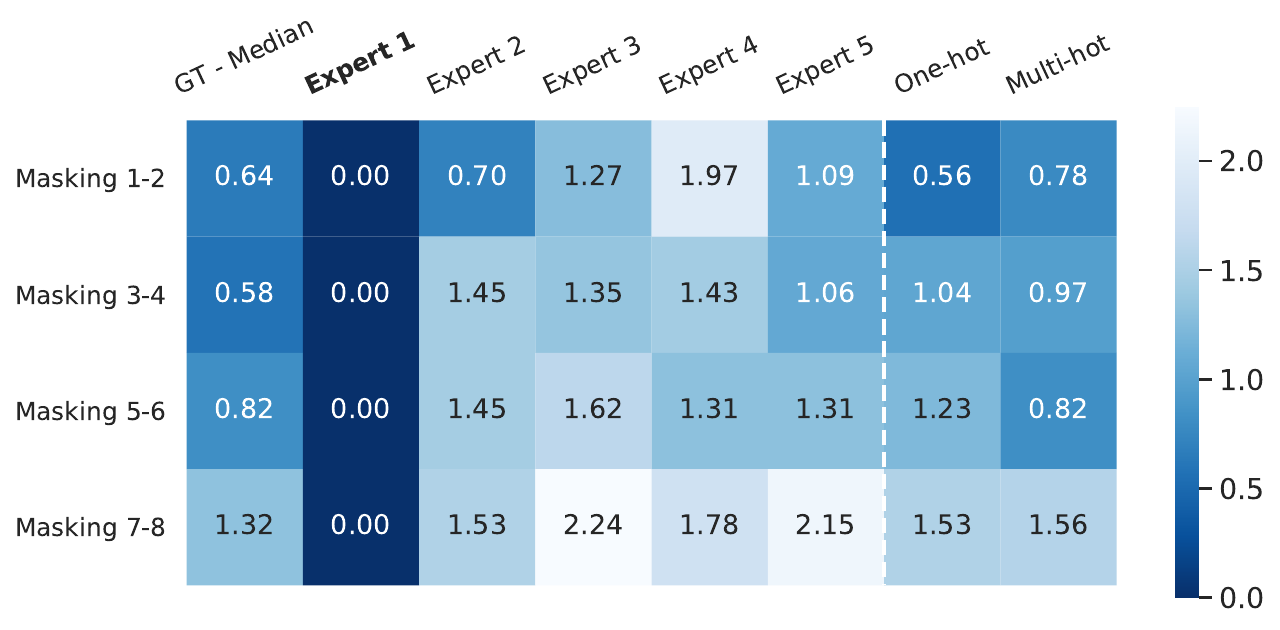}}\quad
\subfloat[{AMAE}, expert 2 as reference \label{fig:amae_4_maskinggrounps_expert2}]{\includegraphics[width=0.45\columnwidth]{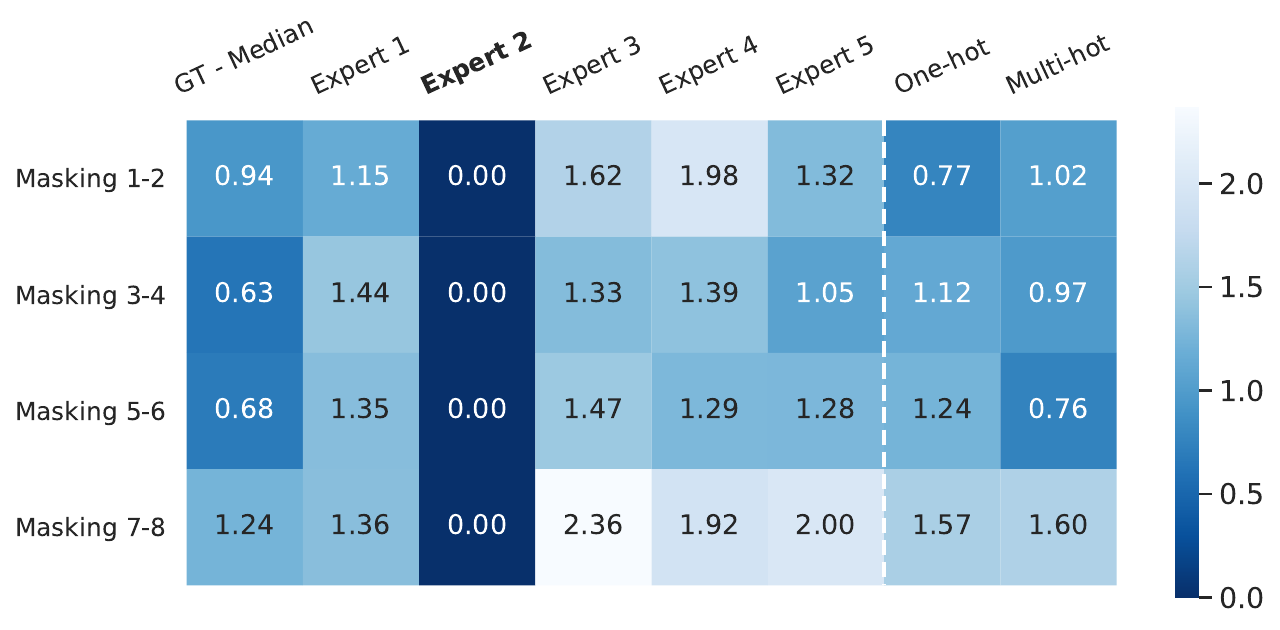}}\quad
\subfloat[{AMAE}, expert 3 as reference
\label{fig:amae_4_maskinggrounps_expert3}]{\includegraphics[width=0.45\columnwidth]{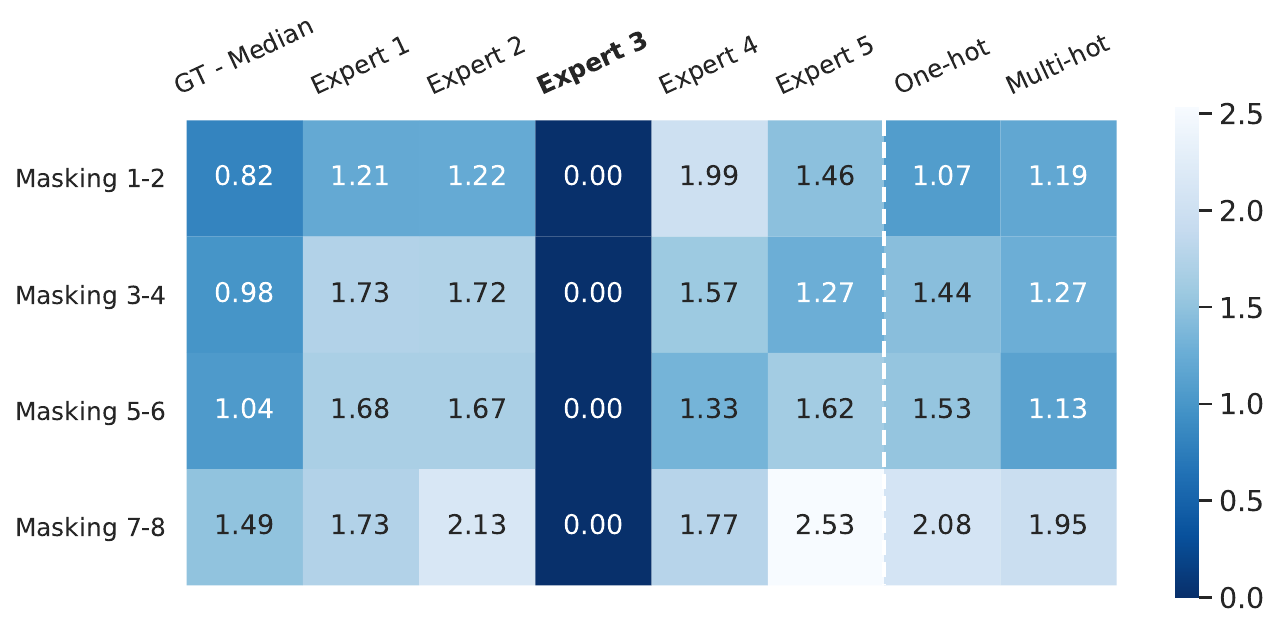}}\quad
\subfloat[{AMAE}, expert 4 as reference \label{fig:amae_4_maskinggrounps_expert4}]{\includegraphics[width=0.45\columnwidth]{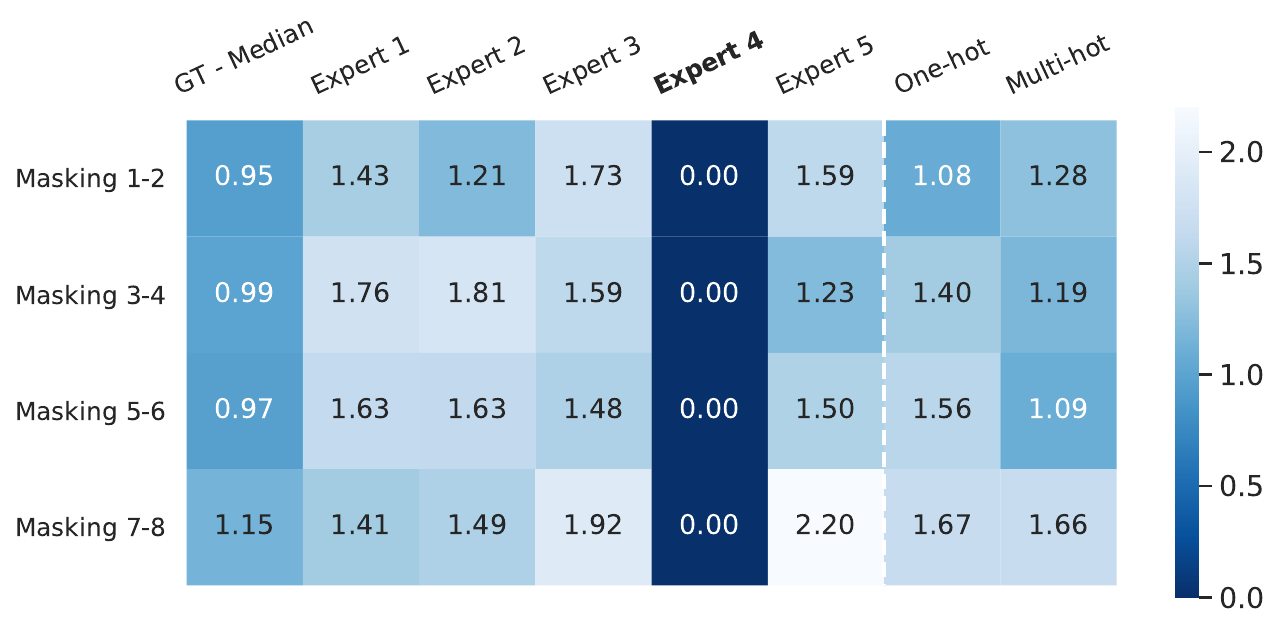}}\quad
\subfloat[{AMAE}, expert 5 as reference
\label{fig:amae_4_maskinggrounps_expert5}]{\includegraphics[width=0.45\columnwidth]{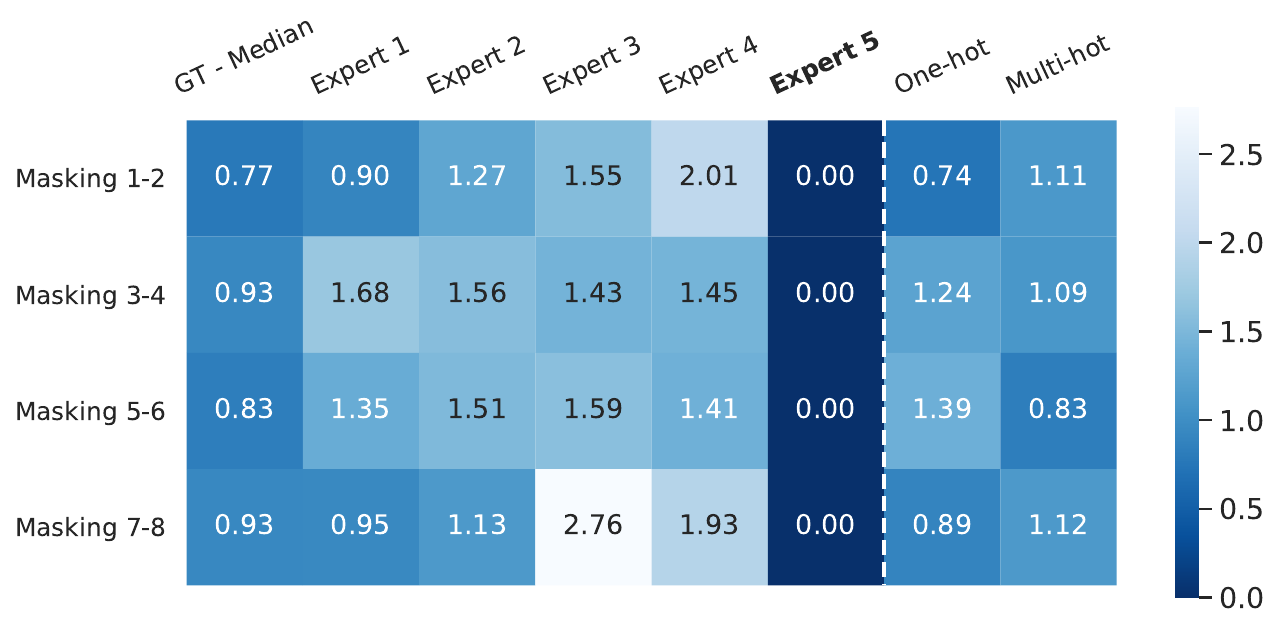}}\quad

\caption{Expert and model agreement in different masking levels on public test set. We perform five runs on our one-hot and multi-hot models and report the mean.} 
\label{fig:amae_4_maskinggrounps}
\end{figure}

In Figure \ref{fig:amae_4_maskinggrounps}, we provide annotator agreements across different masking levels. Here, we divided masking levels into 4 non-overlapping groups: (1, 2), (3, 4), (5, 6), and (7, 8). For each of these groups, we individually considered each expert (and GT-median) as the reference, and evaluated how other experts/models agree with respect to the selected reference in terms of the \textit{AMAE} measure. For example, in the top row of Figure \ref{fig:amae_4_maskinggrounps_gt} we consider the GT-median as reference, and compute the \textit{AMAE} of the other experts. A low value indicates they agree with the GT-median in discriminating masking level 1 from masking level 2. We repeated this process for the other groups to fill in the remaining rows.

From Figure \ref{fig:amae_4_maskinggrounps}, we can observe that in general disagreement is high in higher masking levels. This trend is especially obvious for experts 3, 4 and 5. The experts tended to agree best when discriminating between middle masking levels (3,4) and (5,6). Performance for the low masking levels (1,2) was interesting, as certain experts tended to agree with each other well (experts 1 and 2) while others disagreed strongly. Finally, the trained models also exhibit this trend, and it can be seen that generally the models and the median agree more than other experts.

%For each of these adjacent groups, we individually considered the median and each expert as the reference, and evaluated how other experts/models agree with respect to the selected reference in terms of the AMAE measure in different masking groups. For instance, for levels 1-2, we considered the median as reference and extracted images with labels of 1-2. We then used those images to calculate the AME of other experts/models. We did this for all the adjacent groups mentioned above. We repeated this procedure, each time considering an individual annotator as reference and evaluated other annotators/models with respect to that expert.

%Results for expert and model agreement in different masking levels on public test set are provided in Figure \ref{fig:amae_4_maskinggrounps}.
%One clear observation is that experts 1 and 2 significantly agree with each other and with the median in levels 1-2 (this can be seen in the models as well). Regardless of which expert is chosen as reference, we can see that agreements in levels 1-2 are often higher than other adjacent groups, which was also proven with the  F1-scores in Table~\ref{table:f1}.
%This is also in line with the lower variations in lower masking levels as shown in Figure \ref{fig:violin_pd}. 
%With respect to median, agreements in levels 7-8 are also high for experts 1,2,4. However, with respect to individual annotators, sometimes the agreement in levels 7-8 is not high, an example being the low agreement of expert 3 when either of expert 1 or 2 is chosen as reference.

\section{Dataset accessibility}
%Instructions on how to access the data have been provided in OpenReview along with our submission. 
%Currently, we are actively working on the landing page of the dataset. The public link to the dataset (along with the DOI) will be added to the final version of this paper. 
%We note that the dataset webpage is self-contained, that is, all the data and metadata files needed for the user to understand how to use the data are available in the same place.
Our dataset can be accessed using the DOI: {\datasetDOI}. Access to the dataset files are given upon agreeing to the terms and sending a request. We note that the dataset webpage is self-contained, that is, all the data and metadata files needed for the user to understand how to use the data are available in the same place.

\section{Hosting, licensing, and maintenance plan}\label{sec:hosting}
The dataset is hosted by \url{\SciRepo}. 
This interface has restricted access where users must submit their request, after which the access to the actual files will be granted. 
We have been in contact with staff from SciLifeLab Data Repository and have received a letter from them (available on the final page), supporting the hosting of our data. 
We will actively check for requests made to the data. 
In case any change is required to be made to the dataset, we will use the \textit{versioning} functionality provided by the repository. Through this functionality, all the previous versions of the data will still be available. 
This change will be communicated clearly to the users and will also be reflected on the dataset landing page.

The site contains instructions on how to request the data.
The data is self-contained -- enabling users to easily understand the content and organization of the files using the provided metadata file. Please visit the dataset webpage for license and terms\footnote{The dataset can be found with this DOI: {\datasetDOI}}.

%Our dataset is licensed under \href{https://creativecommons.org/licenses/by-nc-nd/4.0/}{CC BY-NC-ND 4.0} for non-commercial use only. The instructions could be found at: \url{https://creativecommons.org/about/cclicenses/}. Additionally, we will ask users to follow our terms of use when requesting the data, including:
%\begin{enumerate}
%    
%    \item They will not make any attempt to re-identify the screening participants in the dataset or use the data for any illegal purposes, 
    %\item By using CSAW-M, we are given the right to communicate to users in case of change of version of the data, updates etc.
    %\item By using CSAW-M, we are given the right to
%    \item In the unlikely case of re-identification of an individual by a user, the user is asked to immediately let us know,
%    \item They will not share the download link to the data to others,
%    \item If any of the terms of above is violated by a user, the user must delete any copy of the CSAW-M they have.
%\end{enumerate}

\paragraph{Planned service.} SciLifeLab Data Repository will provide an infrastructure to run AI models on the repository using Kubernetes and Docker. Although the plan is not definite yet, we will try to use the provided infrastructure so users could evaluate models trained to estimate masking level using the CSAW-M private test set. This will allow researchers to evaluate their models in a less biased setting, as the (planned) evaluation server will limit the number of times users can submit their models, so models cannot overfit the private test set.

\section{Author statement of responsibility}
%We removed the patient ID's from the data in order to de-identify the individuals in the dataset. To the best of our knowledge, given the available attributes in our dataset, there is no way to identify any subject with regards to the released data.
The authors are responsible for violations of privacy and data ownership laws pertaining to the distribution of CSAW-M, including:
\begin{enumerate}
    \item violations of data protection law, specifically GDPR;
    \item violations of the terms of approvals from the Ethical Review Board (EPM 2021-01030).
\end{enumerate}

The authors bear responsibility for taking the necessary technical and organizational measures to protect the data from re-identification.

%CSAW-M is licensed under \href{https://creativecommons.org/licenses/by-nc-nd/4.0/}{CC BY-NC-ND 4.0} for non-commercial use only. The instructions can be found at: \url{https://creativecommons.org/about/cclicenses/}. Additionally, we ask users to follow terms of use when requesting the data.

\section{Documentation framework and intended use of the dataset}
\label{datasheet}
Here we provide documentation for the CASAW-M dataset. We use the Datasheets for Datasets framework \cite{gebru2018datasheets} to document our dataset.
 %%%%%%%%%%%%%%%%%%%%%%%%%%%%%%%%%%%%%%%%%%%%%%%%%%%%%%%%%%%%%%%%%%%%%%%%%%%%%%%%
%\noindent
% \textbf{
% This document is based on \textit{Datasheets for Datasets} by Gebru \textit{et
% al.} \cite{gebruDatasheetsDatasets2020}. Please see the most updated version
% \underline{\textcolor{blue}{\href{http://arxiv.org/abs/1803.09010}{here}}}.
% }
%%%%%%%%%%%%%%%%%%%%%%%%%%%%%%%%%%%%%%%%%%%%%%%%%%%%%%%%%%%%%%%%%%%%%%%%%%%%%%%%
%%%%%%%%%%%%%%%%%%%%%%%%%%%%%%%%%%%%%%%%%%%%%%%%%%%%%%%%%%%%%%%%%%%%%%%%%%%%%%%%
%\begin{mdframed}

\begin{mdframed}[linecolor=\sectioncolor]
\section*{\textcolor{\sectioncolor}{
    MOTIVATION
}}
\end{mdframed}

\textcolor{\sectioncolor}{\textbf{%
For what purpose was the dataset created?
}
Was there a specific task in mind? Was there
a specific gap that needed to be filled? Please provide a description.
} \\
%%%
The main goal in creating this dataset was to enable the development of models capable of identifying screening participants in mammography screening whose mammograms cannot easily be assessed due to a high level of mammographic masking, a phenomenon that occurs when potential cancer could largely be obscured by the surrounding tissue in the breast. 
As a result, breast cancer in these participants is more likely to be missed during regular mammography. 
More sensitive imaging technologies such as MRI are too costly to be provided for all participants visiting a clinic. 
Due to the large number of mammography images that are taken at clinics, there exists a need to develop an AI model that could help identifying screening participants in higher needs of MRI. 
To develop such an AI model, we noticed a lack of mammographic images containing assessment of masking level directly made by expert radiologists. 
Although public mammographic datasets exist, none of them exactly contains direct potential of masking in mammograms assessed by radiologists. 
Our aim was to fill the gap by collecting a dataset that merely focuses on mammographic masking. 
CSAW-M helps to automate identifying of low- and high-masking mammograms.
 \\
%%% 

\textcolor{\sectioncolor}{\textbf{%
Who created this dataset (e.g., which team, research group) and on behalf
of which entity (e.g., company, institution, organization)?
}
} \\
%%%
The dataset was created in a joint collaboration of researchers from KTH Royal Institute of Technology, Karolinska Institutet, Karolinska University Hospital, and S:t Görans Hospital in Stockholm. \\
%%% 

\textcolor{\sectioncolor}{\textbf{%
Who funded the creation of the dataset?} If there is an associated grant, please provide the name of the grantor and the grant name and number. } \\
%%%
This work was partially supported by MedTechLabs \url{https://www.medtechlabs.se/}, the Swedish Research Council (VR) 2017-04609, and Region Stockholm
HMT 20200958.
%%% 
%Regional Cancer Center Stockholm-Gotland as source for the clinical data

\textcolor{\sectioncolor}{\textbf{%
Any other comments?
}} \\
%%%
None. \\
%%%

%%%%%%%%%%%%%%%%%%%%%%%%%%%%%%%%%%%%%%%%%%%%%%%%%%%%%%%%%%%%%%%%%%%%%%%%%%%%%%%%
\begin{mdframed}[linecolor=\sectioncolor]
\section*{\textcolor{\sectioncolor}{
    COMPOSITION
}}
\end{mdframed}
    \textcolor{\sectioncolor}{\textbf{%
    What do the instances that comprise the dataset represent (e.g., documents,
    photos, people, countries)?
    }
    Are there multiple types of instances (e.g., movies, users, and ratings;
    people and interactions between them; nodes and edges)? Please provide a
    description.
    } \\
    %%%
    The dataset comprises mammographic images, together with metadata that is provided as CSV files. The metadata includes masking potential labels collected from five experts, image acquisition parameters, clinical endpoints \ie cancer attributes and density measures. More details can be found in the main paper.\\
    %%% 
    
    \textcolor{\sectioncolor}{\textbf{%
    How many instances are there in total (of each type, if appropriate)?
    }
    } \\
    %%%
    There are 10,020 screening participants in total, and each participant has 1 mammogram from the MLO view of the breast. 9,523 of the images are in CSAW-M training set with one annotation per image, while the rest 497 images are from a public test set where each image has annotations from 5 experts.\\
    %%% 
    
    \textcolor{\sectioncolor}{\textbf{%
    Does the dataset contain all possible instances or is it a sample (not
    necessarily random) of instances from a larger set?
    }
    If the dataset is a sample, then what is the larger set? Is the sample
    representative of the larger set (e.g., geographic coverage)? If so, please
    describe how this representativeness was validated/verified. If it is not
    representative of the larger set, please describe why not (e.g., to cover a
    more diverse range of instances, because instances were withheld or
    unavailable).
    } \\
    %%%
    CSAW-M is a subset of CSAW, a large population-level cohort of screening mammograms \cite{dembrower2019multi}. We exclude images that are: \textit{a)} from patients with implants;\textit{ b)} biopsy images; \textit{c)} mammograms that are not vendor post-processed; \textit{d)} mammograms with aborted exposure; \textit{e)} mammograms not taken by X-ray photoconductor; \textit{f)} earlier mammograms when there are duplicates in the same exam. We sample screening participants with complete mammography exams taken in Karolinska University Hospital and with Hologic manufacturer. We sampled from the participants according to the procedure mentioned in Section \ref{sec:datacreation} in a way that more mammograms with extreme density values are included (which are more clinically interesting), so the sampled mammograms are not necessarily representative of the larger set. This was done because mammograms in the tails of the percent density distribution (very dense or very fatty) are of the highest clinical interest.\\
    
    %Moreover, in CSAW-M public test set, we randomly sample from healthy screening participants.  Note that images can also be aborted in the annotation process due to missing or declined annoations. For all the reasons described above, CSAW-M does not necessarily result in a distribution that is representative of the original population. 
    % \ms{Explain why it is not representative wrt to the selection criteria we mentioned in the paper}. 
    %\ms{(Do we want to visualize the distributions before and after sampling?)}
    %We include all case patients and randomly select 9\,411 control patients to make evaluation more efficient. \ms{Refer to corresponding Section in the paper for sampling?} \\
    %%% 
    
    \textcolor{\sectioncolor}{\textbf{%
    What data does each instance consist of?
    }
    "Raw” data (e.g., unprocessed text or images) or features? In either case,
    please provide a description.
    } \\
    %%%
     Images in PNG format, certain DICOM acquisition attributes that can be used for preprocessing, clinical endpoints, and masking annotations from 5 experts (as detailed in Table \ref{tab:dataset_summary}). Training images have one annotation while test images have 5 annotations per image.  \\
    %%% 
    
    \textcolor{\sectioncolor}{\textbf{%
    Is there a label or target associated with each instance?
    }
    If so, please provide a description.
    } \\
    %%%
    Yes, the labels are masking levels (from 1-8) of each instance annotated by 5 experts, together with certain clinical endpoints, \ie interval or large invasive cancer.  \\
    %%% 
    
    \textcolor{\sectioncolor}{\textbf{%
    Is any information missing from individual instances?
    }
    If so, please provide a description, explaining why this information is
    missing (e.g., because it was unavailable). This does not include
    intentionally removed information, but might include, e.g., redacted text.
    } \\
    %%%
    No, there is no missing information. The information is complete for all individual instances.  \\
    %%% 
    
    \textcolor{\sectioncolor}{\textbf{%
    Are relationships between individual instances made explicit (e.g., users’
    movie ratings, social network links)?
    }
    If so, please describe how these relationships are made explicit.
    } \\
    %%%
    % Yes, the annotations are made by directly showing mammograms to experts \yl{Moein, I am not sure whether I understand this question right, do you think you could help?} \ms{I don't understand completely either - I think there is no specific relation between instances.}
    Yes, the annotations are explicitly applied to the images, which were shown directly to the experts. 
    \\
    %%% 
    
    \textcolor{\sectioncolor}{\textbf{%
    Are there recommended data splits (e.g., training, development/validation,
    testing)?
    }
    If so, please provide a description of these splits, explaining the
    rationale behind them.
    } \\
    %%%
    Yes, we have recommended training and testing splits. We have benchmarked mammographic masking of cancer on suggested testing splits where there are 5 annotations per image (the median was chosen as ground truth), as opposed to the training set that contains 1 annotation per image. There is no recommended development/validation split. However, we have provided the cross-validation folds that we used when developing the models.\\
    
    %from us but the 5 folds we use for cross-validation in our baseline implementation are clearly stated for reproducibility. \yl{is this correct?} \ms{Yes, although after thinking a bit more I am not sure we want to also release the exact folds} \\
    %%% 
    
    \textcolor{\sectioncolor}{\textbf{%
    Are there any errors, sources of noise, or redundancies in the dataset?
    }
    If so, please provide a description.
    } \\
    %%%
    Yes. Experts may have made wrong button clicks or errors in their comparisons.
    Similarly, there may be clerical errors matching patients with their clinical endpoints.  \\
    %%% 
    
    \textcolor{\sectioncolor}{\textbf{%
    Is the dataset self-contained, or does it link to or otherwise rely on
    external resources (e.g., websites, tweets, other datasets)?
    }
    If it links to or relies on external resources, a) are there guarantees
    that they will exist, and remain constant, over time; b) are there official
    archival versions of the complete dataset (i.e., including the external
    resources as they existed at the time the dataset was created); c) are
    there any restrictions (e.g., licenses, fees) associated with any of the
    external resources that might apply to a future user? Please provide
    descriptions of all external resources and any restrictions associated with
    them, as well as links or other access points, as appropriate.
    } \\
    %%%
    The dataset is hosted by the \url{\SciRepo}. It has restricted access where users must submit their request, after which the access to the actual files could be granted. See Appendix \ref{sec:hosting} for more details. \\
    %%% 
    
    \textcolor{\sectioncolor}{\textbf{%
    Does the dataset contain data that might be considered confidential (e.g.,
    data that is protected by legal privilege or by doctor-patient
    confidentiality, data that includes the content of individuals’ non-public
    communications)?
    }
    If so, please provide a description.
    } \\
    %%%
    Yes, the dataset contains information related to the health status of individuals. The information has been reduced in order not to allow the identification of any individual. In our assessment, the dataset contains only de-identified information.
 \\
    %%% 
    
    \textcolor{\sectioncolor}{\textbf{%
    Does the dataset contain data that, if viewed directly, might be offensive,
    insulting, threatening, or might otherwise cause anxiety?
    }
    If so, please describe why.
    } \\
    %%%
    No. Our dataset mainly contains mammography images, and there is nothing offensive, insulting, or threatening.\\
    %%% 
    
    \textcolor{\sectioncolor}{\textbf{%
    Does the dataset relate to people?
    }
    If not, you may skip the remaining questions in this section.
    } \\
    %%%
    Yes, the source dataset is extracted from a large cohort containing millions of mammograms, collected every 18 to 24 months from screening participants aged 40 to 74 in Stockholm county area. The dataset contains around 10,020 mammograms taken from 10,020 participants. \\
    %%% 
    
    \textcolor{\sectioncolor}{\textbf{%
    Does the dataset identify any subpopulations (e.g., by age, gender)?
    }
    If so, please describe how these subpopulations are identified and
    provide a description of their respective distributions within the dataset.
    } \\
    %%%
    In Sweden, individuals with female personal identity numbers are invited for mammographic screening. Our dataset contains mammogramgraphic screenings from screening participants 40 to 74 years of age. Racial information is not collected in Sweden.\\
    %%% 
    
    \textcolor{\sectioncolor}{\textbf{%
    Is it possible to identify individuals (i.e., one or more natural persons),
    either directly or indirectly (i.e., in combination with other data) from
    the dataset?
    }
    If so, please describe how.
    } \\
    %%%
    No, we have taken appropriate measures to ensure it is not possible. The measures include: (1) we removed all individual identifiers from the data, (2) we down-sampled the mammograms, (3) we removed all unnecessary acquisition attributes --DICOM headers--, (4) we simplified the continuous tumor size attribute to a binary outcome, and (5) we anticipated a gated release mechanism to approve users based on their information and project goals before granting access to the data. Users are also required to explicitly agree not to attempt to de-identify any individuals from the dataset.
\\
    %%% 
    
    \textcolor{\sectioncolor}{\textbf{%
    Does the dataset contain data that might be considered sensitive in any way
    (e.g., data that reveals racial or ethnic origins, sexual orientations,
    religious beliefs, political opinions or union memberships, or locations;
    financial or health data; biometric or genetic data; forms of government
    identification, such as social security numbers; criminal history)?
    }
    If so, please provide a description.
    } \\
    %%%
  The cancer images in our dataset are accompanied with the clinical outcome of the screening corresponding to that image, \ie whether they are diagnosed with interval or large invasive cancer. These attributes, however, are available in our dataset in a binary form and the screening participants are de-identified.  \\
    %%% 
    
    \textcolor{\sectioncolor}{\textbf{%
    Any other comments?
    }} \\
    %%%
    None. \\
    %%%

%%%%%%%%%%%%%%%%%%%%%%%%%%%%%%%%%%%%%%%%%%%%%%%%%%%%%%%%%%%%%%%%%%%%%%%%%%%%%%%%
\begin{mdframed}[linecolor=\sectioncolor]
\section*{\textcolor{\sectioncolor}{
    COLLECTION PROCESS
}}
\end{mdframed}

    \textcolor{\sectioncolor}{\textbf{%
    How was the data associated with each instance acquired?
    }
    Was the data directly observable (e.g., raw text, movie ratings),
    reported by subjects (e.g., survey responses), or indirectly
    inferred/derived from other data (e.g., part-of-speech tags, model-based
    guesses for age or language)? If data was reported by subjects or
    indirectly inferred/derived from other data, was the data
    validated/verified? If so, please describe how.
    } \\
    %%%
    The source images of our dataset are in DICOM format with DICOM metadata. Each patient is linked to the Regional Cancer Registry to define clinical endpoints such as whether a screening participant was healthy or had been diagnosed with breast cancer. Mammograms were shown to five experts to assign masking annotations. \\
    %%% 

    \textcolor{\sectioncolor}{\textbf{%
    What mechanisms or procedures were used to collect the data (e.g., hardware apparatus or sensor, manual human curation, software program, software API)?
    }
   How were these mechanisms or procedures validated?
    } \\
    %%%
    We designed a user interface through which we showed two images alongside each other and asked experts to do pair-wise comparisons and select the image that is harder to assess. It was validated to be bug-free by running several tests with experts before the collection process began. \\
    %%% 
    
    \textcolor{\sectioncolor}{\textbf{%
    If the dataset is a sample from a larger set, what was the sampling strategy (e.g., deterministic, probabilistic with specific sampling probabilities)?
    }
    } \\
    %%%
    The data is sampled from CSAW as explained in Section \ref{sec:datacreation}. The sampling was done in way to include more mammograms with very low or very high percent density measure as these are the most clinically interesting images.  \\
    %%% 
    
    \textcolor{\sectioncolor}{\textbf{%
    Who was involved in the data collection process (e.g., students, crowdworkers, contractors) and how were they compensated (e.g., how much were crowdworkers paid)?
    }
    } \\
    %%%
    Researchers from KTH Royal Institute of Technology, Karolinska Institutet, Karolinska University Hospital, and S:t Görans Hospital in Stockholm were involved in the data collection. All participants were compensated for their time in the course of their normal research activities.\\
    %\yl{we need help with the question about compensation } \ms{If radiologists have to spend some time on research (e.g. weekly), I think we can simply say that was done as part of the research time so it is already compensated} \fs{Fredrik, could you help write how the annotators were compensated?}\\
    %%% 

    \textcolor{\sectioncolor}{\textbf{%
    Over what timeframe was the data collected?
    }
    Does this timeframe match the creation timeframe of the data associated with the instances (e.g., recent crawl of old news articles)? If not, please describe the timeframe in which the data associated with the instances was created.
    } \\
    %%%
    The mammograms were collected during regular mammography screening between 2008 and 2015 at Karolinska University Hospital. The creation of the CSAW-M dataset, including developing the annotation tool, receiving annotations from experts, cleaning data etc. was initiated in June 2020 and lasted until November 2020.
    \\
    %%% 

    % \textcolor{\sectioncolor}{\textbf{
    % What was the resource cost of collecting the data?
    % }
    % (e.g. what were the required computational resources, and the associated
    % financial costs, and energy consumption - estimate the carbon footprint.
    % See Strubell \textit{et al.}\cite{strubellEnergyPolicyConsiderations2019} for approaches in this area.)
    % } \\
    % %%%
    % YOUR ANSWER HERE \\
    % %%% 

    \textcolor{\sectioncolor}{\textbf{%
    Were any ethical review processes conducted (e.g., by an institutional
    review board)?
    }
    If so, please provide a description of these review processes, including
    the outcomes, as well as a link or other access point to any supporting
    documentation.
    } \\
    %%%
    The Regional Ethical Review Board in Stockholm has approved the research. Also, a dedicated agreement between Karolinska Institutet and KTH Royal Institute of Technology has been made to publish the data.
 \\
    %%% 
    
    \textcolor{\sectioncolor}{\textbf{%
    Did you collect the data from the individuals in question directly, or
    obtain it via third parties or other sources (e.g., websites)?
    }
    } \\
    %%%
    The source mammograms were collected by Karolinska University Hospital, the clinical labels were collected by the Regional Cancer Center, and masking labels were collected by showing mammograms to five experts for annotation. \\
    %%% 
    
    \textcolor{\sectioncolor}{\textbf{%
    Were the individuals in question notified about the data collection?
    }
    If so, please describe (or show with screenshots or other information) how
    notice was provided, and provide a link or other access point to, or
    otherwise reproduce, the exact language of the notification itself.
    } \\
    %%%
   No. The need for informed consent was waived by the Ethical Review Board.
 \\
    %%% 
    
    \textcolor{\sectioncolor}{\textbf{%
    Did the individuals in question consent to the collection and use of their
    data?
    }
    If so, please describe (or show with screenshots or other information) how
    consent was requested and provided, and provide a link or other access
    point to, or otherwise reproduce, the exact language to which the
    individuals consented.
    } \\
    %%%
    No. The need for informed consent was waived by the Ethical Review Board.\\
    %%% 
    
    \textcolor{\sectioncolor}{\textbf{%
    If consent was obtained, were the consenting individuals provided with a
    mechanism to revoke their consent in the future or for certain uses?
    }
     If so, please provide a description, as well as a link or other access
     point to the mechanism (if appropriate)
    } \\
    %%%
    Not applicable. \\
    %%% 
    
    \textcolor{\sectioncolor}{\textbf{%
    Has an analysis of the potential impact of the dataset and its use on data
    subjects (e.g., a data protection impact analysis) been conducted?
    }
    If so, please provide a description of this analysis, including the
    outcomes, as well as a link or other access point to any supporting
    documentation.
    } \\
    %%%
    We consulted an expert in GDPR and dealing with personal data from Karolinska Institutet, and our concerns regarding privacy were cleared. \\
    %%% 
    
    \textcolor{\sectioncolor}{\textbf{%
    Any other comments?
    }} \\
    %%%
    None. \\
    %%%

%%%%%%%%%%%%%%%%%%%%%%%%%%%%%%%%%%%%%%%%%%%%%%%%%%%%%%%%%%%%%%%%%%%%%%%%%%%%%%%%
\begin{mdframed}[linecolor=\sectioncolor]
\section*{\textcolor{\sectioncolor}{
    PREPROCESSING / CLEANING / LABELING
}}
\end{mdframed}

    \textcolor{\sectioncolor}{\textbf{%
    Was any preprocessing/cleaning/labeling of the data
    done (e.g., discretization or bucketing, tokenization, part-of-speech
    tagging, SIFT feature extraction, removal of instances, processing of
    missing values)?
    }
    If so, please provide a description. If not, you may skip the remainder of
    the questions in this section.
    } \\
    %%%
    Data preprocessing was done in our baseline implementations. The source images of our dataset were DICOM files whose pixel values we saved as raw PNG images. Using DICOM metadata, we did preprocessing to generate PNG images. Please refer to Section \ref{sec:datacreation} for more details about image preprocessing.
    %we perform a horizontal flip to make all breasts left-posed, rescale the intensity linearly into a proper DICOM window range, and possibly correct inverted contrast images. We apply distance transform to locate mass center and move it to the center of the image. Zero-padding is applied on the images in order to ensure uniform size among images.\\
    %%%

    \textcolor{\sectioncolor}{\textbf{%
    Was the “raw” data saved in addition to the preprocessed/cleaned/labeled
    data (e.g., to support unanticipated future uses)?
    }
    If so, please provide a link or other access point to the “raw” data.
    } \\
    %%%
    The "raw" data was saved as PNG, and we also provide the preprocessing script that was used in our baseline implementation for reproducibility.  \\
    %%%

    \textcolor{\sectioncolor}{\textbf{%
    Is the software used to preprocess/clean/label the instances available?
    }
    If so, please provide a link or other access point.
    } \\
    %%%
    We used Python standard libraries and the preprocessing script is available in the Github repo of the project: {\maskingRepo}.  \\
    %%%

    \textcolor{\sectioncolor}{\textbf{%
    Any other comments?
    }} \\
    %%%
    None. \\
    %%%

%%%%%%%%%%%%%%%%%%%%%%%%%%%%%%%%%%%%%%%%%%%%%%%%%%%%%%%%%%%%%%%%%%%%%%%%%%%%%%%%
\begin{mdframed}[linecolor=\sectioncolor]
\section*{\textcolor{\sectioncolor}{
    USES
}}
\end{mdframed}

    \textcolor{\sectioncolor}{\textbf{%
    Has the dataset been used for any tasks already?
    }
    If so, please provide a description.
    } \\
    %%%
    % Yes. 
    % Together with two other deep models on risk prediction and cancer detection, the masking model trained with our dataset has been used to identify women at high-risk of interval cancer and enable supplemental MRI.
    Yes. The masking model, together with two other models that we developed to perform breast cancer risk prediction and cancer detection, are combined into a single comprehensive model. This clinical workflow is currently implemented at Karolinska University Hospital in a clinical study to help identify screening participants who are most likely to benefit from additional MRI screening. 
\\
    %%%

    \textcolor{\sectioncolor}{\textbf{%
    Is there a repository that links to any or all papers or systems that use the dataset?
    }
    If so, please provide a link or other access point.
    } \\
    %%%
    No. \\
    %%%

    \textcolor{\sectioncolor}{\textbf{%
    What (other) tasks could the dataset be used for?
    }
    } \\
    %%%
    First, our dataset has annotations that are ordinally related and can be used to study ordinal classification or point-wise ranking tasks. Specifically, our public test set contains 5 annotaitons per image, which make it a useful resource to study human noise and bias. Moreover, our dataset which contains more than 10,000 mammograms, is significantly larger than other public mammography datasets (see Table \ref{tab:datasets} in the main paper). It can be used for pretraining deep learning models that would be used in other downstream tasks in a similar domain to mammography images (for more effective transfer learning). And last but certainly not least, we included clinical endpoints as our metadata, making it valuable in clinical studies. We have shown in the paper that our ResNet-34 models trained on estimating masking potential perform better than the breast density counterparts in identifying screening participants diagnosed with interval and large invasive cancers, without being explicitly trained for these tasks. This shows a great promise for the usefulness of our collected labels and motivates developing better models for estimating masking level. \\
    %%%

    \textcolor{\sectioncolor}{\textbf{%
    Is there anything about the composition of the dataset or the way it was
    collected and preprocessed/cleaned/labeled that might impact future uses?
    }
    For example, is there anything that a future user might need to know to
    avoid uses that could result in unfair treatment of individuals or groups
    (e.g., stereotyping, quality of service issues) or other undesirable harms
    (e.g., financial harms, legal risks) If so, please provide a description.
    Is there anything a future user could do to mitigate these undesirable
    harms?
    } \\
    %%%
    We are aware of the fact that biases exist inherently in our data collection, for the following reasons: \textit{(a)} the data was extracted from a certain population, period and region, with certain manufacturers,  \textit{(b)} the annotations were made by radiologists from a certain region, \textit{(c)} we randomly sampled screening participants and intentionally selected breasts that are denser or fattier which resulted in a distribution that is not representative of the real population. We note that clinical studies are crucially required before deploying models in any clinical processes. \\
    %%%

    \textcolor{\sectioncolor}{\textbf{%
    Are there tasks for which the dataset should not be used?
    }
    If so, please provide a description.
    } \\
    %%%
    In the main article, we have noted that our dataset is not aimed for developing/evaluating cancer detection models, as the cancer images in CSAW-M are chosen to be \textit{contralateral} to cancer laterality, \ie the breast that does \emph{not} contain tumor was selected (please refer to Section \ref{sec:datacreation} for motivation).
     \\
    %%%

    \textcolor{\sectioncolor}{\textbf{%
    Any other comments?
    }} \\
    %%%
    None. \\
    %%%

%%%%%%%%%%%%%%%%%%%%%%%%%%%%%%%%%%%%%%%%%%%%%%%%%%%%%%%%%%%%%%%%%%%%%%%%%%%%%%%%
\begin{mdframed}[linecolor=\sectioncolor]
\section*{\textcolor{\sectioncolor}{
    DISTRIBUTION
}}
\end{mdframed}

    \textcolor{\sectioncolor}{\textbf{%
    Will the dataset be distributed to third parties outside of the entity
    (e.g., company, institution, organization) on behalf of which the dataset
    was created?
    }
    If so, please provide a description.
    } \\
    %%%
    \href{\SciRepo}{SciLifeLab Data Repository}, who hosts our dataset, is currently relying on the Figshare service, but plans to move data to its own storage servers soon. This does not change availability of the dataset in any way, nor does it impose additional restrictions by any third party. SciLifeLab Data Repository is affiliated with KTH Royal Institute of Technology and the hosting of our dataset is guaranteed there. \\
    %%%

    \textcolor{\sectioncolor}{\textbf{%
    How will the dataset will be distributed (e.g., tarball on website, API,
    GitHub)?
    }
    Does the dataset have a digital object identifier (DOI)?
    } \\
    %%%
    The dataset webpage could be found with this DOI {\datasetDOI}. All the instructions on how to access the data is clearly mentioned on the dataset landing page, which contains the actual data files along with metadata to help users better understand how to use the data. \\
    %The dataset itself will be available at: \url{\datasetURL}. which is has instructions on how to request access to the files. Once the request is granted, all the files could directly be downloaded. The DOI of our dataset is: \url{\datasetDOI}. \\
    %%%

    \textcolor{\sectioncolor}{\textbf{%
    When will the dataset be distributed?
    }
    } \\
    %%%
    %The dataset is planned to be published on \datasetReleaseDate. Due to the internal preparations
    % The process of publishing the data is planned to initiate on June 8, 2021. However, due to the internal communication between KTH and SciLifeLab, it might take longer until the landing page is set up.
    The dataset has already been distributed with this DOI {\datasetDOI}.
    \\
    %%%

    \textcolor{\sectioncolor}{\textbf{%
    Will the dataset be distributed under a copyright or other intellectual
    property (IP) license, and/or under applicable terms of use (ToU)?
    }
    If so, please describe this license and/or ToU, and provide a link or other
    access point to, or otherwise reproduce, any relevant licensing terms or
    ToU, as well as any fees associated with these restrictions.
    } \\
    %%%
    %Our dataset is licensed under \href{https://creativecommons.org/licenses/by-nc-nd/4.0/}{CC BY-NC-ND 4.0} for non-commercial use only. The instructions could be found at: \url{https://creativecommons.org/about/cclicenses/}. Additionally, we will ask users to follow terms of use when requesting the data. Please refer to Appendix \ref{sec:hosting} for the details of the ToU.
    Yes. Please visit the dataset home page for details about the license and terms. 
    %The dataset webpage can be found with this DOI {\datasetDOI}.
  \\
    %%%

    \textcolor{\sectioncolor}{\textbf{%
    Have any third parties imposed IP-based or other restrictions on the data
    associated with the instances?
    }
    If so, please describe these restrictions, and provide a link or other
    access point to, or otherwise reproduce, any relevant licensing terms, as
    well as any fees associated with these restrictions.
    } \\
    %%%
    There is no third parties imposed IP-based or other restrictions on the data associated with the instances. %There are no restrictions imposed by any third parties. The use of the data must follow what was approved by the Ethical Review Board, i.e. data may be shared with other research groups who perform research of relevance to our research questions which are: exploring potential improvements in breast cancer screening, exploring potential improvements in preoperative staging and prediction of prognosis and treatment response - based on image analysis and other risk factors. \yl{Moein, I need help}
\\
    %%%

    \textcolor{\sectioncolor}{\textbf{%
    Do any export controls or other regulatory restrictions apply to the
    dataset or to individual instances?
    }
    If so, please describe these restrictions, and provide a link or other
    access point to, or otherwise reproduce, any supporting documentation.
    } \\
    %%%
    There are no export controls or other regulatory restrictions on this dataset to the best of our knowledge. \\
    %%%

    \textcolor{\sectioncolor}{\textbf{%
    Any other comments?
    }} \\
    %%%
    None. \\
    %%%

%%%%%%%%%%%%%%%%%%%%%%%%%%%%%%%%%%%%%%%%%%%%%%%%%%%%%%%%%%%%%%%%%%%%%%%%%%%%%%%%
\begin{mdframed}[linecolor=\sectioncolor]
\section*{\textcolor{\sectioncolor}{
    MAINTENANCE
}}
\end{mdframed}

    \textcolor{\sectioncolor}{\textbf{%
    Who is supporting/hosting/maintaining the dataset?
    }
    } \\
    %%%
    The  data  is  currently  supported/hosted by \url{\SciRepo} (the support letter could be seen on the final page of this article). The infrastructure for hosting and maintaining the data is guaranteed to be supported  by the repository. \\
    %%%

    \textcolor{\sectioncolor}{\textbf{%
    How can the owner/curator/manager of the dataset be contacted (e.g., email
    address)?
    }
    } \\
    %%%
    The owners of the dataset could be contacted through either of the following email addresses: \url{yue3@kth.se} and \url{sorkhei@kth.se}.
    \\
    %%%

    \textcolor{\sectioncolor}{\textbf{%
    Is there an erratum?
    }
    If so, please provide a link or other access point.
    } \\
    %%%
    There is no erratum for the dataset. \\
    %%%

    \textcolor{\sectioncolor}{\textbf{%
    Will the dataset be updated (e.g., to correct labeling errors, add new
    instances, delete instances)?
    }
    If so, please describe how often, by whom, and how updates will be
    communicated to users (e.g., mailing list, GitHub)?
    } \\
    %%%
    At the moment, there is no plan for any updates. In case the dataset is updated, the most recent version of it could be seen on the dataset website (previous versions will still be visible), and the DOI will also change accordingly with respect to the version. \\
    %%%

    \textcolor{\sectioncolor}{\textbf{%
    If the dataset relates to people, are there applicable limits on the
    retention of the data associated with the instances (e.g., were individuals
    in question told that their data would be retained for a fixed period of
    time and then deleted)?
    }
    If so, please describe these limits and explain how they will be enforced.
    } \\
    %%%
    In case of retention, the data will be deleted and a new version that addresses the issue will be re-uploaded, in which case we ask users to delete their old copy of data (our ToU covers this). This is communicated clearly to the users. \\
    %%%

    \textcolor{\sectioncolor}{\textbf{%
    Will older versions of the dataset continue to be
    supported/hosted/maintained?
    }
    If so, please describe how. If not, please describe how its obsolescence
    will be communicated to users.
    } \\
    %%%
    If there is a change in the version of the dataset, previous versions will still be hosted and supported on the website. We will announce the change of version as explicit as possible on the website. \\
    %%%

    \textcolor{\sectioncolor}{\textbf{%
    If others want to extend/augment/build on/contribute to the dataset, is
    there a mechanism for them to do so?
    }
    If so, please provide a description. Will these contributions be
    validated/verified? If so, please describe how. If not, why not? Is there a
    process for communicating/distributing these contributions to other users?
    If so, please provide a description.
    } \\
    %%%
    We always welcome if experts in the area of mammography are interested in contributing to our dataset by assessing mammographic potential masking of tumor in our mammograms. Code for our annotation tool with complete instructions on how to use it is publicly available at \url{https://github.com/MoeinSorkhei/CSAW-M_Annotation_Tool/}. For further discussion, experts are very welcome to contact us using the contact info on the website. We will then compare the received annotations against our ground truth using the same metrics we used in the paper. Finally, the annotations and the comparison against our ground-truth will be made publicly available on our website, acknowledging the contribution. We are also interested in receiving BI-RADS annotations. We would be happy to discuss any other possible contributions not mentioned here. We note, however, that although contributions will be made publicly visible on our website, they do not result in any change in the authors of the dataset.
    \\
    %%%

    \textcolor{\sectioncolor}{\textbf{%
    Any other comments?
    }} \\
    %%%
    None. \\
    %%% % import the content of the datasheet from sections folder

% letter of support
\begin{figure}
    \centering
    \includegraphics[scale=0.7]{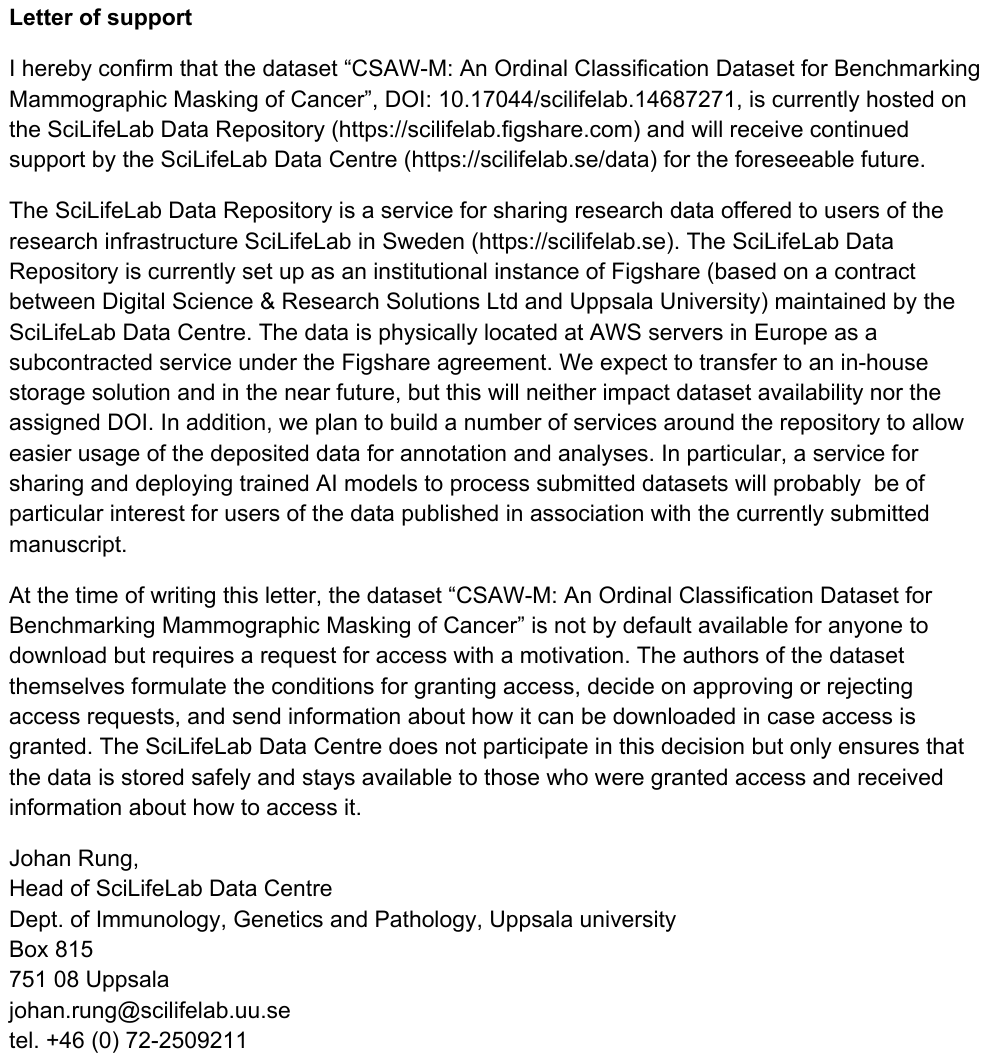}
    %\caption{}
    \label{fig:letter}
\end{figure}

\end{document}